\documentclass[10pt,journal,compsoc]{IEEEtran}
\ifCLASSOPTIONcompsoc
  \usepackage[nocompress]{cite}
\else
  \usepackage{cite}
\fi
\ifCLASSINFOpdf
\else
\fi

\usepackage{amsmath}
\usepackage{algorithmic}
\usepackage{graphicx}
\usepackage{textcomp}
\usepackage{xcolor}
\usepackage{booktabs}
\usepackage[utf8]{inputenc}
\usepackage{tikz}
\usepackage{subfigure}
\usepackage{amssymb}

\begin{document}
%
\title{Use of Affective Visual Information \\ for Summarization of Human-Centric Videos}
%
%
%
%

\author{Berkay Köprü,
        Engin Erzin

\thanks{B. Köprü is with the KUIS AI Lab and Electrical \& Electronics Engineering Department, Ko\c{c} University, Istanbul, Turkey. \protect\\
E. Erzin is with the KUIS AI Lab, Computer Engineering Department and Electrical \& Electronics Engineering Department, Ko\c{c} University, Istanbul, Turkey. \protect\\
E-mail: \{bkopru17, eerzin\}@ku.edu.tr
}
}

\IEEEtitleabstractindextext{%
\begin{abstract}
Increasing volume of user-generated human-centric video content and their applications, such as video retrieval and browsing, require compact representations that are addressed by the video summarization literature. Current supervised studies formulate video summarization as a sequence-to-sequence learning problem and the existing solutions often neglect the surge of human-centric view, which inherently contains affective content. In this study, we investigate the affective-information enriched supervised video summarization task for human-centric videos. First, we train a visual input-driven state-of-the-art continuous emotion recognition model (CER-NET) on the RECOLA dataset to estimate emotional attributes. Then, we integrate the estimated emotional attributes and the high-level representations from the CER-NET with the visual information to define the proposed affective video summarization architectures (AVSUM). In addition, we investigate the use of attention to improve the AVSUM architectures      and propose two new architectures based on temporal attention (TA-AVSUM) and spatial attention (SA-AVSUM). We conduct video summarization experiments on the TvSum database. The proposed AVSUM-GRU architecture with an early fusion of high level GRU embeddings and the temporal attention based TA-AVSUM architecture attain competitive video summarization performances by bringing strong performance improvements for the human-centric videos compared to the state-of-the-art in terms of F-score and self-defined face recall metrics. 
\end{abstract}

}

\maketitle

\IEEEraisesectionheading{\section{Introduction}\label{sec:introduction}}

Multimedia applications and services have seen a surge in recent years. There is more than 500 hours upload per minute to video streaming platforms such as Youtube and Twitch. Especially user-generated human-centric videos allocate major part of the available videos in these video streaming platforms. Such a massive video resource creates two crucial needs: (i) personalized human-computer interactions (HCI), and (ii) efficient representations and retrieval of video contents. Emotion recognition addresses the former by providing affect aware applications \cite{kopruCERMTL2020}, while video summarization addresses the latter \cite{videoSumSurvey1}.

Emotion recognition is a critical task that enables personalized HCI applications and understanding of personal choices. Humans are emotional creatures, even for many cognitive tasks decision making processes are driven primarily by emotions \cite{emotionDecisionMaking2015}. Emotions are represented both in discrete and continuous domains. Discrete categorical emotions, such as happiness and sadness, can also be represented in the 3-dimensional continuous affect space of activation, valence, and dominance attributes, which are the indicators of activeness-passiveness, positiveness-negativeness, and dominance-submissiveness, respectively \cite{threeDimensionalEmotion12,threeDimensionalEmotioin54}.

In the literature, estimation of activation, valence and dominance attributes is referred as Continuous Emotion Recognition (CER). Although CER is widely studied by the speech processing community \cite{speechCER16,speechCER17,speechCER19}, it has been studied over the visual channels \cite{videoCER1,videoCER2,videoCER3}. In \cite{videoCER1}, a deep attention based convolutional network is proposed to detect facial expressions and emotions. Aspandi et al. \cite{videoCER2} proposed an adversarial training approach to jointly estimate whether the image is fake or not while estimating the activation and valence (AV) attributes. In \cite{videoCER3}, VGG-16 driven visual features are used as the input of a stacked convolutional recurrent neural network (CRNN) for affect recognition in the wild. 

Compact and efficient representations of videos can be extracted by selecting subsets of frames from the videos through video summarization techniques. Video summarization is categorized regarding the output type which could be video frames or video fragments \cite{videoSumSurvey1,frameimportanceToGroundTruth2016,earlyVideoSum1}. The former, finding key frames, is known as video storyboard generation. The latter,  chronologically stitched segment selection, is referred as video skimming. Due to the discrete nature of storyboard generation, it lacks in smoothness and naturalness. However, since it does not require synchronization, it offers more degrees of freedom in data organization than video skimming \cite{videoSumSurvey1}.

Early video summarization approaches are unsupervised and made use of low level similarity measures between the frames \cite{earlyVideoSum5,earlyVideoSum6,earlyVideoSum2ColorHist,earlVideoSum3ColorHist,earlyVideoSum4MI}; while recent unsupervised studies apply Generative Adversarial Networks (GANs) \cite{unsupervisedVideoSum1} and attention \cite{unsupervisedVideoSumAttention1,unsupervisedVideoSumAttention2} to solve the video summarization problem. They use GANs to reconstruct the input video from the selected key frames for unsupervised video summarization. Also low-level measures, such as color histogram intersections \cite{earlyVideoSum2ColorHist,earlVideoSum3ColorHist} and mutual information \cite{earlyVideoSum4MI}, are investigated for unsupervised video summarization.

Supervised approaches for video summarization have recently become popular that tend to perform better in other computer vision tasks such as object detection and segmentation. In \cite{frameimportanceToGroundTruth2016,videoSumFCSN2018,videoSummAttentionEncoderDecoder2020,supervisedVideoSumRW1}, the authors model the video summarization task as a sequence-to-sequence mapping problem. In \cite{frameimportanceToGroundTruth2016}, Long Short-Term Memory (LSTM) is adopted to tackle with variable range dependencies within the encoder-decoder type architectures. In contrast to recurrent models, \cite{videoSumFCSN2018} proposes a fully convolutional solution, where all frames are processed together. Recent approaches \cite{videoSummAttentionEncoderDecoder2020,supervisedVideoSumRW1}, succesfully adapted attention mechanisms into the video-summarization task. 

Coupling of affective computing and video summarization has not been extensively studied in the literature. In two related video summarization studies, key frames are selected based on the physiological responses of the viewers that are expected to be highly correlated with the viewers' emotional states \cite{affectiveSumFromViewer1,affectiveSumFromViewer3}. In \cite{affectiveSumFromViewer1}, the facial activity of the viewer is tracked, and then the frames are ranked to extract personal highlights from the videos by using heuristics. Money and Agius track physiological responses of the viewers, such as heart rate and electrodermal response, to analyze sub-segments of the videos \cite{affectiveSumFromViewer3}. These studies leverage physiological information that are highly related to emotional states of the humans to generate personal highlights/summaries. On the other hand, these approaches require at least one subject as a video viewer and do not provide feasible solutions for the automatic video summarization.

In this study, we address the use of affective information, which is extracted from visual data, for the summarization of human-centric videos. Unlike studies evaluating viewers' perspective \cite{affectiveSumFromViewer1,affectiveSumFromViewer3}, we model affective states from humans in the video. That makes automatic affective video summarization feasible and specific for human-centric videos. For affective video summarization, we adopt a two-step approach. First, affective information is extracted using the convolutional recurrent neural networks. Then, we explore the affective information and attention mechanisms to enrich a fully convolutional network based video summarization. To summarize, the main contributions of this study are as follows:
\begin{itemize}
    \item We formulate a novel end-to-end learning problem for affective-information enriched video summarization targeting human-centric videos.
    \item We model affective information in terms of emotional attributes and learned embeddings extracted from the CER module. 
    \item We investigate the use of attention mechanisms in video summarization for human-centric videos and develop temporal and spatial attention-based frameworks.
    \item We carry out affective video summarization evaluations on the state-of-the-art using both standard and self-defined evaluation metrics.
\end{itemize}

The rest of this paper is organized as follows. Section~2 reviews related work, and Section~3 describes main building blocks of the proposed framework. Section~4 presents the experiments conducted together with the performance evaluations. Finally, conclusion is presented in Section~5.

\section{Related Work}

Our proposed affective video summarization framework integrates emotion recognition into video summarization. In this regard, we first point out several emotion recognition studies that relates video summarization. Then, we briefly discuss unsupervised and supervised video summarization literature.

Emotion recognition studies formulate the problem as discrete emotion recognition (DER) \cite{discreteER1,discreteER2,discreteER3} or continuous emotional attribute regression \cite{contER_CCC1,kopruCERMTL2020}. In \cite{discreteER1}, stacked CNN-RNN architecture is proposed to extract local and global features to classify emotions. For the CER problems, Schmitt et al. investigates RNNs with CCC loss function after extracting low-level descriptors (LLDs) such as mel-frequency cepstral coefficients (MFCCs) and zero crossing rate from speech signal \cite{contER_CCC1}. Recently, multi-modal approaches are explored for CER and DER \cite{multimodalCERRW1,kopruCERMTL2020}. Tzirakis et al. design an end-to-end network utilizing raw video, audio and text, where visual features are extracted using 3 stage High-Resolution Network and audio features are extracted via multiple 1-D convolutional layers \cite{multimodalCERRW1}. Contextual features are extracted from text by first generating point-wise n-grams using convolutional layers, then linearly projecting these sub-features with multiple heads to increase the diversity. Finally, extracted features are fused using attention.  In \cite{kopruCERMTL2020}, visual information is expressed using facial attributes and audio information is represented by the MFCCs. Audio-visual information is fused at the feature level, and a CRNN model is trained with multi-task learning for the CER problem. 

Two recent emotion recognition studies are interesting in the context of video summarization \cite{emotionSummarization1,emotionSummarization2}.
Xu et al. investigates information transfer from image and textual data of videos for emotion recognition, emotion attribution and emotion-oriented summarization \cite{emotionSummarization1}. First, they learn video representations using Image Transfer Encoding and textual representations using zero-shot learning from auxiliary datasets. Then, they perform a categorical emotion recognition using the Support Vector Machine (SVM) classifier. Later, emotion attribution sets the contribution of each frame to the video’s overall emotion. Finally, video summarization is formulated as a selection of key frames by maximizing an emotion attribute based score function.
In the second related study, Tu et al. train a joint model to capture emotion attribution and recognition using multitask learning \cite{emotionSummarization2}. Later, similar to the first study \cite{emotionSummarization1}, video summarization task is formulated as a post-processing optimization problem and solved using MINMAX dynamic programming. 
Note that both of these studies formulate summarization task as a optimization problem which is executed in post-processing. Hence their video summarization frameworks are not learning based and they do not explore how affective information alter behavior of the proposed summarization architecture. Furthermore, the proposed solutions are not evaluated on the state-of-the-art video summarization datasets.

Scarcity of the labeled data leads unsupervised learning studies for the video summarization problem.
In a recent study, Jung et al. address unsupervised video summarization problem by first learning  discriminative features over a Variational Auto Encoder (VAE) and GAN-based architecture using variance loss to alleviate ineffective feature learning \cite{unsupervisedVideoSumRW1}. Then, they define a chunk and stride network (CSNet) to overcome the difficulty of learning for long-length videos.
In another study, Zhao et al. presents a dual learning framework for the unsupervised video summarization \cite{unsupervisedVideoSumRW3}. They integrate the summary generation and video reconstruction tasks using multi-task learning so to reward the summary generator under the assistance of the video reconstructor.
Zhou et al. formulates summarization task as sequential decision-making using an end-to-end reinforcement learning based framework \cite{unsupervisedVideoSumRW2}. They utilize a reward function that jointly accounts for diversity and representativeness of the generated summaries in an unsupervised setting. 

Supervised studies on video summarization adapts encoder-decoder architectures \cite{videoSumFCSN2018,videoSummAttentionEncoderDecoder2020, encodeDecoderVSum2019, supervisedVideoSumRW1}.  In \cite{encodeDecoderVSum2019}, video is modeled as a 3-dimensional tensor, and 3D convolutional networks are used in the encoder to extract shallow and deep spatio-temporal features. Then, extracted multi-level features are fed into an LSTM based decoder. The proposed architecture is trained with the Sobolev loss, which constrains the derivative of the sequential data. With the great success of attention modules \cite{attention1,attention2}, recent works in video summarization adapt attention into the decoder part. Ji et al. extract visual features using the GoogleNet and later encode with a Bidirectional LSTM network \cite{videoSummAttentionEncoderDecoder2020}. Encoded vectors combine Bahdanau attention \cite{attention1}, and then feed into an LSTM based decoder. Later the Bidirectional LSTM approach is extended to prevent semantic information loss \cite{videoSummAttentionEncoderDecoder2021}. In the extended architecture, an additional network analyzing the semantic information loss is added and used as a feedback mechanism from the decoder to the encoder using the Huber loss. 

Although we apply supervised learning for video summarization task, our study differs from these studies in the exploitation of affective information for the video summarization task.
\section{Methodology} \label{sec:method}
In this paper, we investigate the use of affective information for enriching video summarization by capturing emotionally salient regions of the human-centric videos. First, we state and formulate the video summarization problem and define the visual feature extraction for both the CER and video summarization tasks. Then, an end-to-end framework for the CER is presented. Emotional attributes and high-level embeddings from the CER framework are later used as affective representations by the video summarization framework. Finally, we introduce the proposed affective video summarization architectures by first defining a video summarization baseline and then enriching this baseline with the fusion of affective information.

\begin{figure}
    \centering
    \includegraphics[width=\columnwidth]{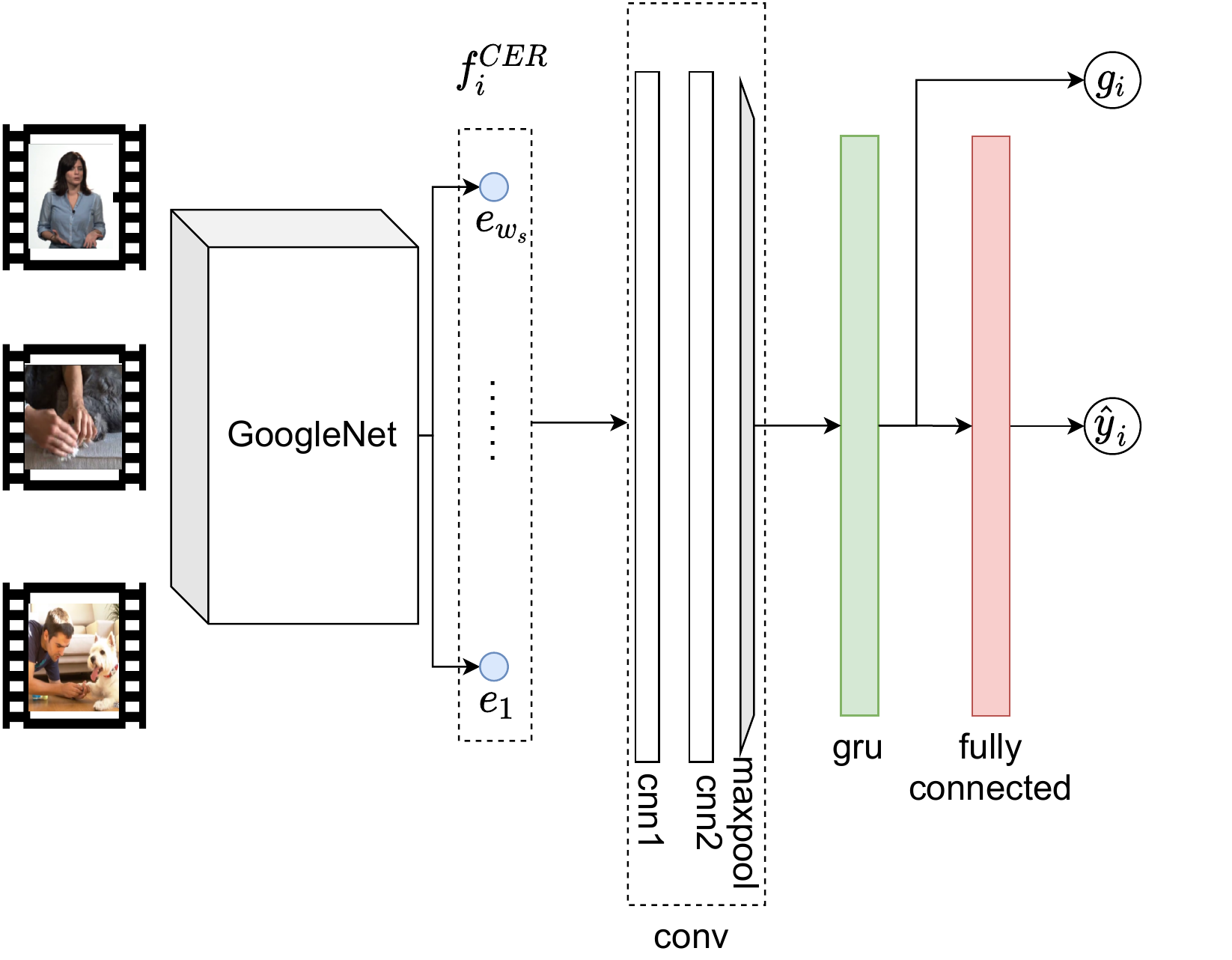}
	\caption{CER-NET: The continuous emotion recognition network
	}
	\label{fig:cer-net}
\end{figure}

\begin{figure*}
    \centering
    \includegraphics[width=\textwidth]{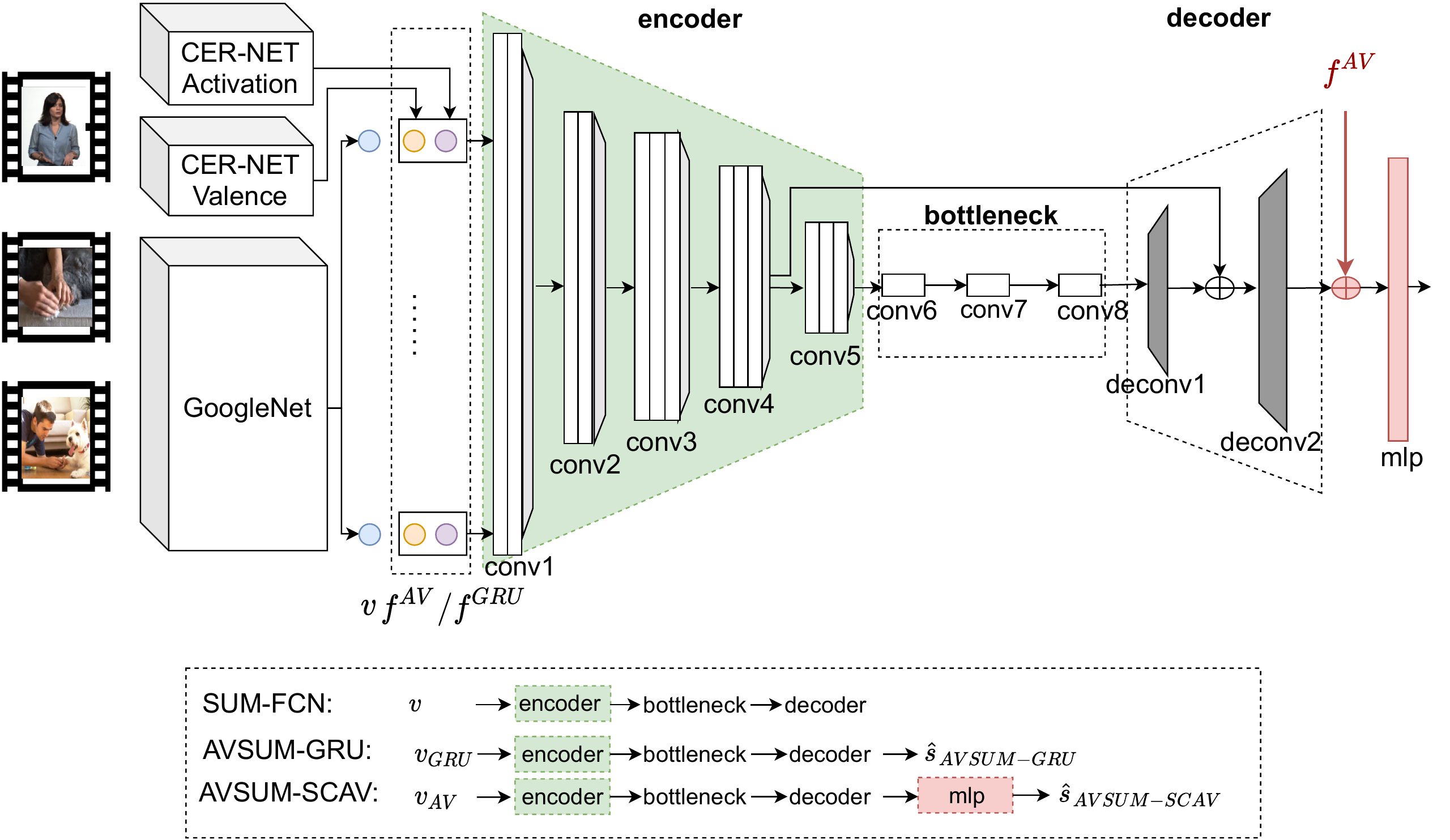}
    \caption{An overview of the proposed AVSUM-GRU and AVSUM-SCAV architectures. AVSUM-GRU combines high-level affective information $f_{\text{GRU}}$ from the CER-NET estimators with the GoogleNet drived visual features. AVSUM-SCAV defines a skip-connection for the emotional attributes $f_{\text{AV}}$ as well as combining them with the GoogleNet drived visual features for the video summarization. 
    }
    \label{fig:avsum}
\end{figure*}

\subsection{Problem Statement}
\label{subsec:prblm}
Video summarization is widely formulated as either a binary classification or a frame-level regression task. In the binary classification task, summarization outputs are either key-frames \cite{videoSumFCSN2018,frameimportanceToGroundTruth2016} or key-shots \cite{tvSum2018,frameimportanceToGroundTruth2016} from the video. On the other hand, frame-level importance scores are extracted in the regression task \cite{videoSummAttentionEncoderDecoder2020,frameimportanceToGroundTruth2016}.

In this study, we formulate the video summarization as a binary classification task where the positive labels correspond to the selected key-frames. The summarization network receives a feature matrix, $\mathbf{v} \in \mathbb{R}^{N \times D}$, and emits an output matrix, $\mathbf{s} \in \mathbb{R}^{N \times C}$, where $N$ is the number of frames in the video, $D$ is the dimensionality of the frame-level visual feature, and $C$ is the number of classes. We take $C=2$ representing the positive and negative classes for the key-frame selection and these two nodes output class probability values at the output of the network. Then the positive class for key-frame selection or the negative class for frame skip are set by picking the node with higher probability. Eventually, the video summary is constructed from the key-frames that are labeled as positive.

In this study, we extract the visual information using the {\it GoogleNet} \cite{googlenet2015}. Output of the \textit{pool5} layer of the pre-trained {\it GoogleNet} is used as the visual feature and represented as $\mathbf{e}_i \in \mathbb{R}^{D}$ at frame $i$ with dimension $D=1024$.

\subsection{CER Network (CER-NET)}

The continuous emotion recognition problem is set as the continuous regression of an emotional attribute from the temporal visual features. For this purpose, we construct a CER network (CER-NET), which consists of two back-to-back convolutional layers, a max pool layer in the temporal domain, a Gated Recurrent Unit (GRU) layer, and a fully connected layer as shown in Figure~\ref{fig:cer-net}. We use the CER-NET to train two separate networks to estimate the activation and valence (AV) attributes separately. A temporal visual feature matrix is defined to be the input of the CER-NET. For this purpose, visual features around the $i$-th frame are cascaded to define the temporal visual feature as
\begin{equation}
\mathbf{f}^{\text{CER}}_i = [\mathbf{e}_{i-\Delta+1},\ldots,\mathbf{e}_{i-1},\mathbf{e}_{i},\mathbf{e}_{i+1},\ldots,\mathbf{e}_{i+\Delta}] \in \mathbb{R}^{D \times T}
\end{equation}
at frame $i$, where $T$ is the temporal window size and it is set as $T = 2\Delta = 20$~frames. 

We refer the group of back-to-back two convolutional and a max-pool layers as the \textit{conv} layer for the sake of simplicity. 
In CER-NET, the \textit{conv} layer models the spatial relations and provides a compact representation of the temporal window. In this manner, both temporally related features are highlighted and dimensionality of the representation is reduced. Dimensionality reduction is the key to complexity reduction and it also prevents overfitting. The compact representation from the \textit{conv} layer is fed into GRU to model long-term temporal relations. 

At the inference phase, CER-NET receives $\mathbf{f}^{\text{CER}}_i$ and provides the GRU layer outputs as $\mathbf{g}_i \in  \mathbb{R}^{G}$ and the fully connected layer outputs as $\hat{y}_i \in \mathbb{R}$. We define the GRU layer output, $\mathbf{g}_i$, as an affective embedding, which can carry affective information to the video summarization model. On the other hand, the fully connected layer output, $\hat{y}_i$, represents the estimated emotional attribute (A or V) at frame~$i$ and delivers another source of affective information.

\subsection{Affective Video Summarization} \label{ss:sum_nets}

The proposed affective-information enriched video summarization (AVSUM) is based on a fully convolutional neural network (FCN) for semantic segmentation \cite{fcn2014}, which is adapted by \cite{videoSumFCSN2018} into video summarization (SUM-FCN). In the following, we briefly describe SUM-FCN and then present two fusion architectures to combine affective information with video summarization, which are later extended by temporal and spatial attention mechanisms.

\subsubsection{SUM-FCN}

SUM-FCN adopts an encoder-decoder architecture where the encoder is based on fully convolutional layers and the decoder is based on deconvolutions. Figure~\ref{fig:avsum} includes the architecture of the SUM-FCN compromising encoder-bottleneck-decoder layers driven by the visual input $\mathbf{v}$. The encoder compresses the temporal information while increasing spatially, and the bottleneck passes it to the decoder to reconstruct necessary information from this compact representation. SUM-FCN receives the visual features for the whole video at once and provides the summarization outputs at once. Video frame rate is typically down-sampled before the summarization. Let $\mathbf{v}$ be the input of SUM-FCN, then $\mathbf{v} = [\mathbf{e}_{1},\ldots,\mathbf{e}_{N}]^{'} \in \mathbb{R}^{N \times D}$ where $N$ is number of frames in the down-sampled stream. Given $\mathbf{v}$, SUM-FCN emits $\mathbf{s} \in \mathbb{R}^{N \times C}$, where $C$ is the dimension of the summarization annotations and set as $C=2$ as defined in section~\ref{subsec:prblm}.

\subsubsection{Affective Feature Fusion for AVSUM}

Affective information is extracted from the two CER-NET models, which are trained to estimate the activation and valence (AV) attributes separately. Two types of affective information cues are extracted from the CER-NET models: (i) the estimated AV attributes ($\mathbf{f}^{\text{AV}}$), and (ii) the learned high-level CER-NET representations based on the GRU embeddings ($\mathbf{g}^{\text{AV}}$). The estimated AV attribute vector $\mathbf{f}^{\text{AV}}_{j}$ is constructed as a column vector from the estimated activation and valence attributes as 
$\mathbf{f}^{AV}_{j} = [\hat{y}_{j}^A,\hat{y}_{j}^V]^{'} \in \mathbb{R}^{2}$ at frame $j$ in the down-sampled stream. Similarly, $\mathbf{g}^{\text{AV}}_{j}$ is constructed by concatenating the outputs of the GRU layers from the two CER-NET models as $\mathbf{g}^{\text{AV}}_{j}= [{\mathbf{g}_{j}^{\text{A}}}{'},\mathbf{g}_{j}^{\text{V}}{'}]^{'} \in \mathbb{R}^{2G}$. Then, the emotional attribute $\mathbf{f}^{\text{AV}}$ and the affect embedding $\mathbf{f}^{\text{GRU}}$ representations of the video are defined as 
\begin{align}
\label{eq:fAV}
    \mathbf{f}^{\text{AV}} &= [\mathbf{f}^{\text{AV}}_{1}, \ldots, \mathbf{f}^{\text{AV}}_{N}]^{'} \in \mathbb{R}^{N \times 2} \\
    \label{eq:fGRU}
    \mathbf{f}^{\text{GRU}} &= [\mathbf{g}^{\text{AV}}_{1}, \ldots, \mathbf{g}^{\text{AV}}_{N}]^{'} \in \mathbb{R}^{N \times 2G}.
\end{align}

The first proposed AVSUM architecture, referred as AVSUM-GRU, combines the affect embedding $\mathbf{f}^{\text{GRU}}$ with the visual input $\mathbf{v}$. Figure~\ref{fig:avsum} presents the AVSUM-GRU architecture receiving the $\mathbf{v}_{\text{GRU}}$ input as
\begin{align}
    \mathbf{v}_{\text{GRU}} &= \mathbf{v} \oplus \mathbf{f}^{\text{GRU}} \in \mathbb{R}^{N \times (D+2G)},
    \label{eq:affective_features2}
\end{align}
where $\oplus$ operator is representing the feature vector combining over the whole video. 

Alternatively, we define the AVSUM-SCAV architecture, as illustrated in Figure~\ref{fig:avsum}, by combining the emotional attribute  $\mathbf{f}^{\text{AV}}$ with the visual input $\mathbf{v}$ to receive $\mathbf{v}_{AV}$ as
\begin{align}
    \mathbf{v}_{\text{AV}} &= \mathbf{v} \oplus \mathbf{f}^{\text{AV}} \in \mathbb{R}^{N \times (D+2)}
    \label{eq:affective_features} 
\end{align}
and incorporating a long skip-connection by concatenating the emotional attribute $\mathbf{f}^{\text{AV}}$ to the final layer of the summarization network. AVSUM-SCAV inserts a skip-connection to the output of \textit{deconv2} layer and has a final fully connected \textit{mlp} layer to reduce the dimension from $C+2$ to $C$.

\subsubsection{Temporal Attention for AVSUM}

\begin{figure*}[t]
    \centering
    \includegraphics[width=\textwidth]{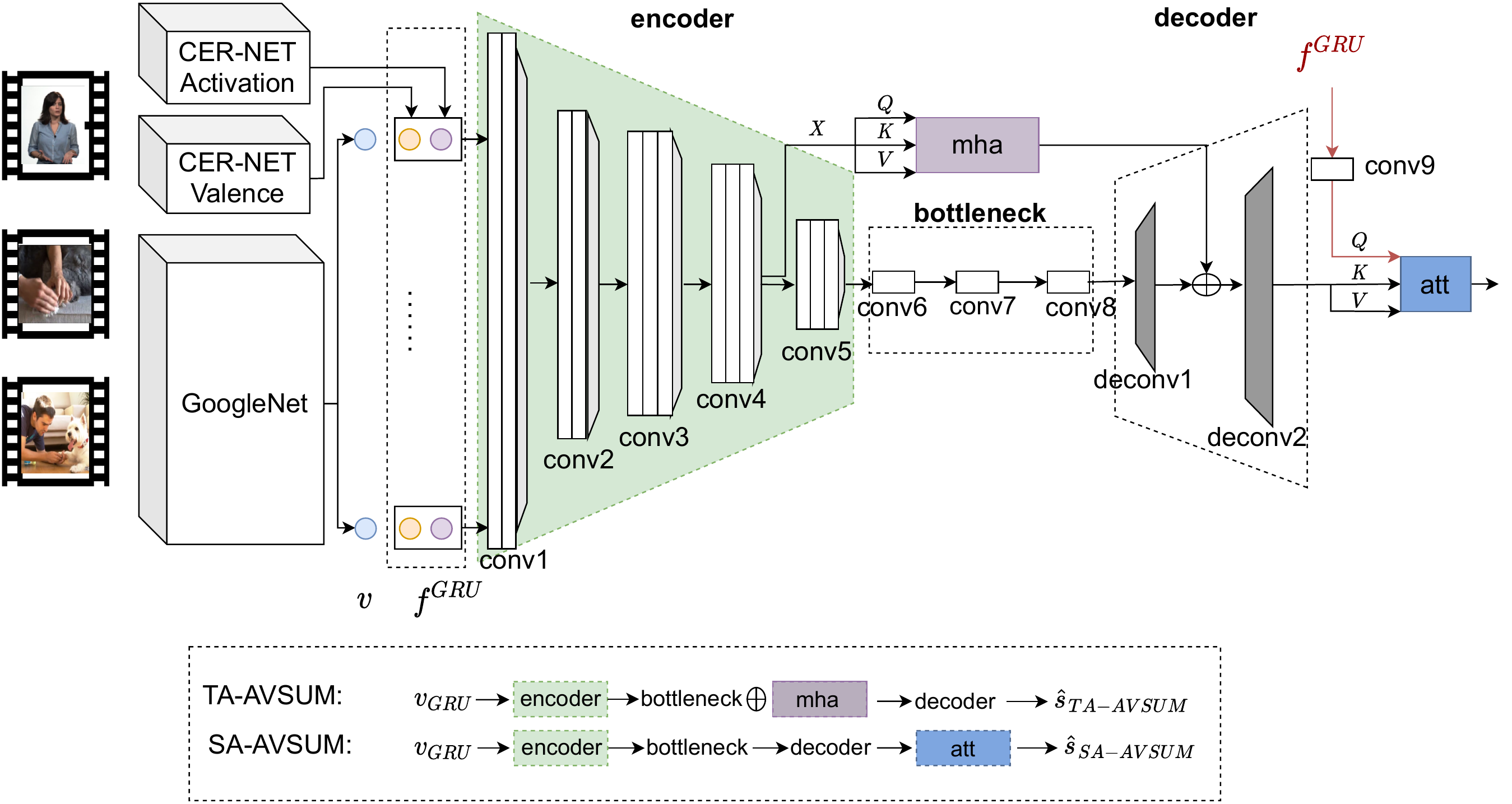}
    \caption{An overview of the proposed TA-AVSUM and SA-AVSUM architectures. TA-AVSUM applies multi-head-self attention to the output of \textit{conv4} layer ($X$), on top of AVSUM architecture.  On the other hand, SA-AVSUM modifies the long skip connection of affective features by applying a spatial attention.
    }
    \label{fig:ta_sa_avsum}
\end{figure*}

We adapt multi-headed attention (MHA) mechanism into the AVSUM architecture to efficiently model temporal dependencies across the video frames. Figure~\ref{fig:ta_sa_avsum} depicts the MHA based temporal attention structure, referred as TA-AVSUM, where MHA is placed to the output of \textit{conv4} layer which emits $\mathbf{X} \in \mathbb{R}^{M\times S}$. Here, $M$ and $S$ are respectively the temporal and spatial dimensions of $\mathbf{X}$.

MHA receives three inputs as Query ($\mathbf{Q}$), Key ($\mathbf{K}$) and Value ($\mathbf{V}$), then outputs a weighted summation of the rows of $\mathbf{V}$. The weights are calculated from the similarity between the $\mathbf{Q}$ and $\mathbf{K}$. In this context, the MHA is defined as \begin{align}
    \mathrm{head}_h = \text{softmax}(\frac{\mathbf{Q}\mathbf{W}_h^{Q}(\mathbf{K}\mathbf{W}_h^K)^T}{\sqrt{M}})\mathbf{V}\mathbf{W}_h^V \label{eq::single_attention}\\
    \mathrm{MHA}(\mathbf{Q},\mathbf{K},\mathbf{V}) = \mathrm{Concat}(\mathrm{head}_1, \ldots ,\mathrm{head}_H) \mathbf{W}^{O},
\end{align}
where $\mathbf{W}_h^Q$, $\mathbf{W}_h^K$, $\mathbf{W}_h^V$ and $\mathbf{W}^O$ are the learned linear projections and $H$ is the number of heads. We employ multi-headed self-attention, by setting $\mathbf{Q}$, $\mathbf{K}$ and $\mathbf{V}$ to $\mathbf{X}$. Hence, the learned projections matrices are formed as $\mathbf{W}_h^Q, \; \mathbf{W}_h^K, \; \mathbf{W}_h^Q \in \mathbb{R}^{S \times \frac{S}{H}}$, and $\mathbf{W}^O \in \mathbb{R}^{M \times M}$.

\subsubsection{Spatial Attention for AVSUM}
Motivated by the Squeeze and Excitation Networks \cite{squeexeExcitationNet}, which attend to different channels of an image, we propose the fourth AVSUM network with spatial domain attention and refer it as SA-AVSUM. Figure~\ref{fig:ta_sa_avsum} depicts the SA-AVSUM structure where we employ attention to the output of the \textit{deconv2} layer. Unlike TA-AVSUM, we adopt single-headed attention, which is formulated in (\ref{eq::single_attention}). Then, $\mathbf{K}$ and $\mathbf{V}$ are set to transpose of $\hat{\mathbf{s}}_{\text{AVSUM}}$ and $\mathbf{Q}$ is set from the affective embedding $\mathbf{f}^{\text{GRU}}$.

\subsection{Model Training}

Training of the CER-NET and video summarization models are executed in two phases. First the CER-NET models are trained, later they are fixed and integrated for the affective video summarization to train the AVSUM-GRU, AVSUM-SCAV, TA-AVSUM, and SA-AVSUM models.

\subsubsection{CER-NET} 
The CER-NET models are trained separately to estimate the activation and valence attributes using the concordance correlation coefficient (CCC) based loss function. The loss function of the CER-NET is defined as the negated CCC value,
\begin{equation}
    L_{\text{CER}} = -\frac{2\sigma_{y\hat{y}}^{2}}{\sigma_{y}^2+\sigma_{\hat{y}}^2+(\mu_{y}-\mu_{\hat{y}})^2}, \label{eq:ccc}
\end{equation}
where $y$ is the ground truth and $\hat{y}$ is the estimated attribute.

\subsubsection{Video Summarization Networks}

The key frame selection problem has an imbalanced nature, since only a small number of frames are selected for the summary \cite{videoSumFCSN2018}. In order to overcome the imbalance problem, a weighted binary cross entropy loss is defined as \begin{equation}
    L_{\text{SUM}} = -\frac{1}{N}\sum^{N}_{j=1} w_{z_{j}}(z_{j}\log{\hat{z_j}}+(1-z_{j})\log{(1-\hat{z_{j}}))},
\end{equation}
where $z_j \in {0,1}$ is the binary ground truth, $\hat{z}$ is the predicted score and $w_{z_{j}}$ is the weight of the $j^{\text{th}}$ frame. The weights for the binary target are defined as
\begin{equation}
    w_{0} = \frac{1}{N}\sum_{j=1}^N z_j \;\;\;  \mathrm{and} \;\;\;
    w_{1} = 1 - w_{0}.
\end{equation}


\section{Experiments} \label{sec:experiments}

Experimental evaluations of the proposed models and comparisons with the state-of-the-art are performed using two datasets. In this section, we first introduce the datasets and evaluation metrics, then explain implementation details. Finally, experimental results are presented and discussed.

\subsection{Datasets}
\label{sec:databases}
We train and evaluate the CER-NET on the RECOLA dataset, which is a popular multi-modal dataset for emotion recognition \cite{recola2013}. The RECOLA dataset is composed of multi-modal recordings of dyadic conversations from 27 French speakers. From these 27 recordings, 18 of them are annotated, and the rest of the records are used for testing. The annotations are at the rate of 40~msec and from 6~different annotators. In total, we use 90 minutes of recordings from the RECOLA dataset in this study.

\begin{figure}[!hbt]
    \centering
    \includegraphics[width=\columnwidth]{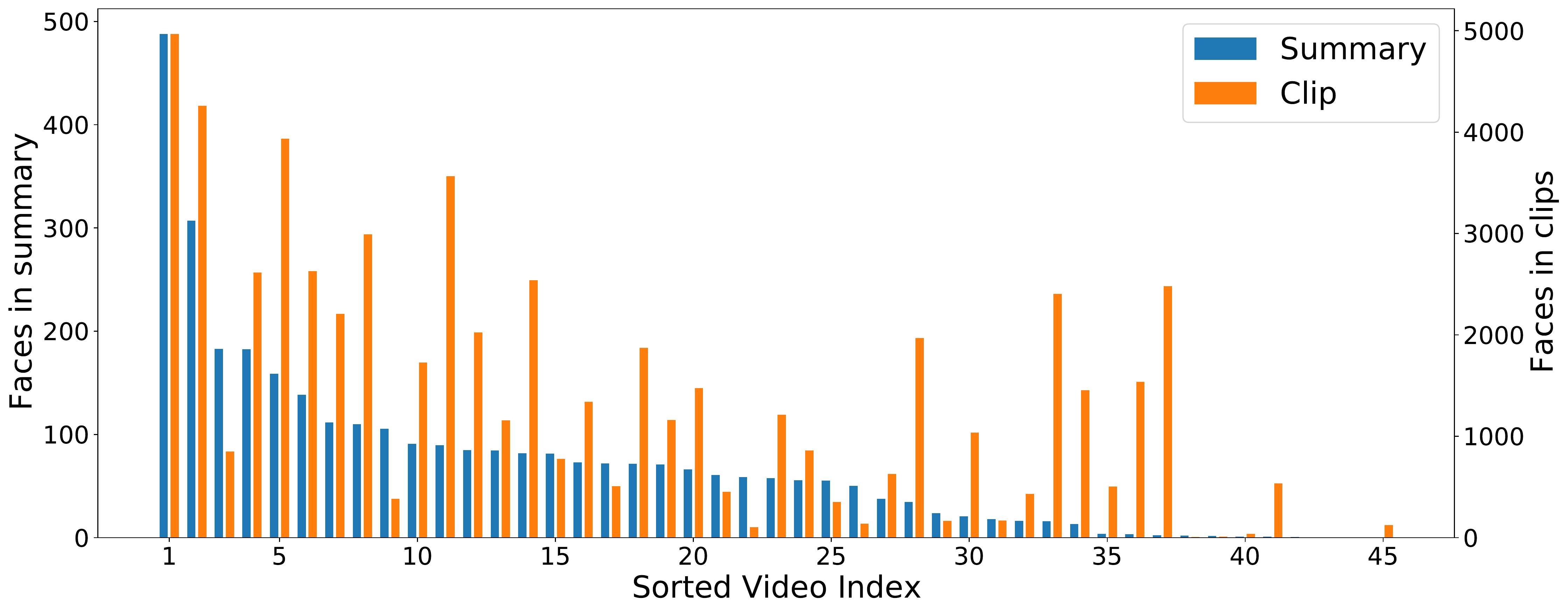}
	\caption{Number of face including frames in the summary and total video for TvSum dataset where the videos are sorted regarding the number of faces in the summary.
	}
	\label{fig:no_face_tvsum}
\end{figure}

Experimental evaluations on the proposed AVSUM architectures are executed on the frequently used TvSum dataset \cite{tvSum2018}. Note that there is no available video summarization dataset containing only human-centric videos in the literature. The TvSum dataset contains 50 user generated videos from 10 different categories, such as vehicle tire changing, sandwich making, grooming an animal, etc. The demographics of the dataset in terms of  face including frames in the summary and in total video clip is presented in Figure~\ref{fig:no_face_tvsum}. We categorize videos into human-centric and rest regarding the number of faces in the summary where we set the threshold to 80 frames labeling 15 videos as human-centric. Ground truths are provided by the frame-level importance score for each video from 20 raters. We follow the approach in \cite{frameimportanceToGroundTruth2016, videoSumFCSN2018} to convert the frame-level importance scores into the keyshot-based summaries.

\subsection{Evaluation Metrics}

Emotional attribute estimation with the CER-NET is evaluated using the CCC metric following \cite{kopruCERMTL2020}, which is the negative of the $L_{\text{CER}}$ loss in (\ref{eq:ccc}). 

For the video summarization task, following \cite{videoSumFCSN2018, frameimportanceToGroundTruth2016, videoSummAttentionEncoderDecoder2020}, F-score is used as the evaluation metric. For the $k$-th video, let the ground truth binary summarization vector be $\mathbf{s}^k$ and estimated binary summarization vector to be $\mathbf{\hat{s}}^k$ where $\mathbf{s}^k, \mathbf{\hat{s}}^k \in \mathbf{R}^{N}$. Then precision $P^k$ and recall $R^k$ for the $k$-th video are calculated as
\begin{equation}
    P^{k} = \frac{\mathbf{s}^{k} \cdot \mathbf{\hat{s}}^{k}}{\sum_{j}^{N}\hat{s}_j^{k}} \;\;\; \mathrm{and} \;\;\; 
    R^{k} = \frac{\mathbf{s}^{k} \cdot \mathbf{\hat{s}}^{k}}{\sum_{j}^{N}s_j^{k}}, 
\end{equation}
where $j$ runs over the frames. Video level F-score is defined as the harmonic mean of the precision and recall,
\begin{align}
    F1^{k} &= \frac{2P^kR^k}{P^k+R^k}.
\end{align}
Then, the final F-score metric $F1$ is defined as the unweighted average of the video level F-score values as
\begin{align}
    F1 &= \frac{1}{K} \sum_{k=1}^{K} F1^{k},
\end{align}
where $K$ is the number of videos in the dataset.

Since the affective information is learned from a dataset of human-centric videos, capability of capturing affective frames of the video during the summarization is expected to be better with human-centric videos.
By highlighting this fact, we define two new metrics for the evaluation of the affective video summarization. The first metric computes normalized F1 score differences with the baseline over the videos with the highest number of face appearances. To define this metric, let us first define the normalized F1 score difference for the $k$-th video with respect to the SUM-FCN baseline model as
\begin{align}
    \Delta F1 ^{k}& = \frac{F1^{k} - F1^{k}_{SUM-FCN}}{F1^{k}_{SUM-FCN}}, 
\end{align}
where $F1^k$ refers to the F1 score of the model in evaluation. Also assume that all the videos in the dataset are sorted with the highest number of face appearances in descending order and are indexed with $k_l$ for $l=1,\dots,K$. Then, the cumulative F1 score difference metric for the \textit{Top-L} human-centric videos is defined as
\begin{align}
    \Delta{F1}_L &=\sum_{l=1}^{L} \Delta F1^{k_l}.
\end{align}
Associated with the $\Delta{F1}_L$, we also compute the F1 score of the \textit{Top-L} human-centric videos and refer it as $F1_L$. 


Human-centric nature of the videos can be associated with the face appearances in the video frames. Motivated with this fact, we set a second metric for evaluation of the affective video summarization as the recall rate of face appearing frames in the extracted summary. 
Let $\mathbf{d}_{F}^{k}$ be the binary vector representing whether a frame includes a face appearance or not for the $k$-th video. Binary face appearance vectors are extracted by the histogram of oriented gradients (HOG) based face detector \cite{hog_face_detector_dlib}. Then, the recall rate of face appearing frames, let's refer it as face recall, $R$ is defined as
\begin{align}
    R &= \frac{1}{K} \sum_{k=1}^{K} \frac{\mathbf{\hat{s}}^{k} \cdot (\mathbf{d}_F^{k} \odot \mathbf{s}^{k}) }{\sum_{j}^{N}s_j^{k}},
\end{align}
where $(\mathbf{d}_F^{k} \odot \mathbf{s}^{k})$ is the Hadamard product of $\mathbf{d}_{F}^{k}$ and $\mathbf{s}^{k}$.

\begin{table}[b]
\begin{center}
\caption{CCC performance of the CER-NET emotion recognition model on the RECOLA dataset}
\label{tab:recola_performance}
\begin{tabular}{l c c c}
\toprule[1pt]\midrule[0.3pt]
\textbf{Model}                             & \textbf{Activation} & \textbf{Valence} \\ \midrule \hline
{CER-NET}                                  &{0.40}              &{0.17} \\ \hline
{CER-MTL Facial \cite{kopruCERMTL2020}}    &{0.15}              & {0.06} \\ \hline
{End-to-end Visual Network \cite{endEndVisualNetCER}} &{0.36}   &{0.48} \\ \hline
\end{tabular}
\end{center}
\end{table}

\begin{table*}[!ht]
\begin{center}
\caption{Cumulative F-score ($F1$) and face recall ($R$) together with the \textit{Top-15} F-score ($F1_{15}$) and face recall ($R_{15}$) performances (top two models are in bold) of the AVSUM and the SUM-FCN models with the maximization of $F1$ and $R$ model selection criteria} \label{tab:f1_fd_res}
\begin{tabular}{l|c c c c |c c c c}
\toprule[1pt]\midrule[0.3pt]
\textbf{Model} &\multicolumn{4}{c|}{\textbf{Max} $F1$} & \multicolumn{4}{c}{\textbf{Max} $R$}   \\ 
    & $F1$ (\%) & $F1_{15}$ (\%) & $R$ (\%) &  $R_{15}$ (\%)& $F1$ (\%) & $F1_{15}$ (\%) & $R$ (\%)  & $R_{15}$ (\%)\\ \midrule \hline
\textbf{SUM-FCN \cite{videoSumFCSN2018}}      
& 57.46 & 60.00  & 53.20 & \textbf{65.52} & \textbf{54.10} & 57.40 & 60.62 & 60.28   \\ \hline
\textbf{AVSUM-GRU} 
& \textbf{57.50} & \textbf{60.25}  & \textbf{54.04} & 59.14 & \textbf{54.02} & 57.40 & 60.77 & \textbf{66.20} \\ \hline
\textbf{AVSUM-SCAV}                           
& 56.64 & \textbf{60.60} & 52.11 & 63.80 & 53.80 & 57.42 & \textbf{62.06} & 59.64   \\ \hline
\textbf{TA-AVSUM}                         
& \textbf{57.47} & 59.95  & \textbf{53.22} & \textbf{65.55} & 53.79 & \textbf{59.23} & \textbf{65.12} & \textbf{70.31} \\ \hline
\textbf{SA-AVSUM}                         
& 55.92 & 59.98  & 49.25 & 62.38 & 52.76 & \textbf{59.90} & 58.85 & 62.55 \\ \hline
\end{tabular}
\end{center}
\end{table*}

We also study statistical differences of a given affective feature dimension, $f$, across face appearing and not-appearing frames. For this purpose, Kullback-Leibler (KL) divergence of $f$ in these two classes is defined as
\begin{equation}
    D(P_{f}||Q_{f}) = \sum_{x \in \mathcal{X}} P_{f}(x) \log(\frac{ P_{f}(x)}{ Q_{f}(x)}),
\end{equation}
where $P_{f}$ and $Q_{f}$ are respectively probability distributions of the affective feature dimension $f$ across face appearing and face not-appearing frames. Note that the affective feature dimension $f$ is driven from the affective feature set as $f \in \{f^{A}, f^{V}, g^{AV}_1, g^{AV}_2, \dots, g^{AV}_{2G}\}$.

\subsection{Implementation Details}

\subsubsection{CER-NET} 
The window size $T$ is selected as 20 frames. The \textit{cnn1} and \textit{cnn2} layers have 10 filters, max-pooling layer reduces the temporal dimension from 20 to 5, \textit{gru} layer has 10 cells and \textit{FCN} layer has 1 node. Hence, the dimensionality of $\mathbf{f}^{\text{AV}}$ is 2 and $\mathbf{g}_{\text{AV}}$ is 20 ($G=10$).

Annotated recordings of the RECOLA dataset are used during the training of the CER-NET. RECOLA recordings are divided as 10\% for the test, 10\% for the validation, and 80\% for the training. We applied Adam optimizer with a learning rate of $10^{-4}$ and with the batch size of 256.   

\subsubsection{Video Summarization Networks} Following \cite{videoSumFCSN2018, frameimportanceToGroundTruth2016}, TvSum videos are downsampled to 2~fps, and frames are fed into GoogleNet. For the AVSUM-SCAV, the input feature dimension is set as 1026. The input dimension of the AVSUM-GRU, TA-AVSUM, and SA-AVSUM models became 1044 with the GRU embeddings. For the TA-AVSUM, we set the number of heads $H=4$.

We mimic fixed size cropping in semantic segmentation by uniformly sampling the video frames and using $N=320$ \cite{videoSumFCSN2018}. We adopted the leave one-group out cross-validation technique to compare the performances. At each fold, 9 videos are selected, and the rest of the videos are used for training. The training is held for 50 epochs with a batch size of 5 videos. During the training phase, Adam optimizer with the learning rate of $10^{-3}$ is used. For each training fold, a model achieving the highest F-score ($F1$), and a model achieving the highest face recall ($R$) are selected for the performance evaluations.

\subsection{CER-NET Performance}
Table~\ref{tab:recola_performance} presents the CCC performance of the CER-NET emotion recognition model on the RECOLA dataset. The proposed architecture performs better at estimating the activation than the valence. The end-to-end CER-NET architecture outperforms CER-MTL Facial model \cite{kopruCERMTL2020} in both activation and valence by achieving CCC of 0.40 and 0.17 respectively. The end-to-end visual network in \cite{endEndVisualNetCER} performs strongly for the valence estimation and fairly close with the CER-NET for the activation estimation. CER-NET outperforms \cite{endEndVisualNetCER} by 4\% at estimating the activation, while \cite{endEndVisualNetCER} achieves CCC of 0.48 for valence and CER-NET achieves 0.17. Different than CER-NET, CER-MTL Facial receives visual information as facial activation units and optical flow vectors. Due to the input dimension, CER-NET has more trainable coefficients leading to a more complex structure than the CER-MTL Facial. 

In comparison with these two baseline visual models \cite{kopruCERMTL2020,endEndVisualNetCER}, CER-NET performs competitively in representing affective information on the visual channel. 

\subsection{Cumulative AVSUM Performance}
In this section, we first present performance evaluations of the proposed affective video summarization models in terms of cumulative F-score and face recall metrics. Then, the video level performances are investigated to better highlight characteristics of videos that have improved summarization performance with the affective cues.

Table~\ref{tab:f1_fd_res} presents cumulative F-score ($F1$) and face recall ($R$) together with the \textit{Top-15} F-score ($F1_{15}$) and face recall ($R_{15}$) performances for the proposed AVSUM and the baseline SUM-FCN models. In each column, top-two scoring performances are highlighted in bold. Recall that we apply two model selection criteria based on F-score and face recall maximization. Each model selection criterion is observed to favor its related performance metric in the evaluations. That is, maximization of $F1$ (Max~$F1$) yields higher F-score while maximization of $R$ (Max~$R$) yields higher face recall $R$.

Observing the cumulative $F1$ performances, AVSUM-GRU model is competitive for both model selection criteria and observed as the best performing model with the Max~$F1$ criterion. On the other hand, the cumulative $R$ score highlights AVSUM-GRU and the temporal attention based TA-AVSUM models with the Max~$F1$ criterion and AVSUM-SCAV and TA-AVSUM models with the Max~$R$ criterion. The temporal attention based TA-AVSUM model especially performs significantly better with the Max~$R$ criterion achieving 65.12\% face recall rate.

Affective information is modeled with the continuous emotion recognition task trained on the RECOLA dataset. Since RECOLA is an human-centric dataset, which includes facial videos, we also choose to evaluate the video summarization performance for the \textit{Top-L} human-centric videos, where $L$ is set as 15 with the discussion in section~\ref{sec:databases}. Table~\ref{tab:f1_fd_res} presents the \textit{Top-15} F-score $F1_{15}$ and face recall $R_{15}$ performances. While attention based SA-AVSUM and TA-AVSUM models perform best for the $F1_{15}$ score with the Max~$R$ criterion, AVSUM-SCAV and AVSUM-GRU models perform best with the Max~$F1$ criterion. Overall, the best performance is 60.60\% $F1_{15}$ score with the Max~$F1$ criterion for the AVSUM-SCAV model. Observing the  \textit{Top-15} face recall $R_{15}$ performances, while TA-AVSUM model is competitive with the baseline SUM-FCN model with the Max~$F1$ criterion, it performs significantly superior with the Max~$R$ criterion achieving 70.31\% face recall $R_{15}$ rate.

Table~\ref{tab:f1_fd_res} highlights two runner up models, AVSUM-GRU and TA-AVSUM. AVSUM-GRU model sustains strong F-score rates with the Max~$F1$ criterion, especially for human-centric videos targeted with $F1_{15}$ performance. Alternatively, while temporal attention based TA-AVSUM model performs strongly for the face recall with the Max~$R$ criterion and attains 70.31\% face recall $R_{15}$ rate, it also sustains a competitive performance for the F-score and face recall rates with the Max~$F1$ criterion. Hence temporal attention is observed to better integrate the affective information for the affective video summarization.

\begin{figure}
    \centering
    \includegraphics[width=\columnwidth]{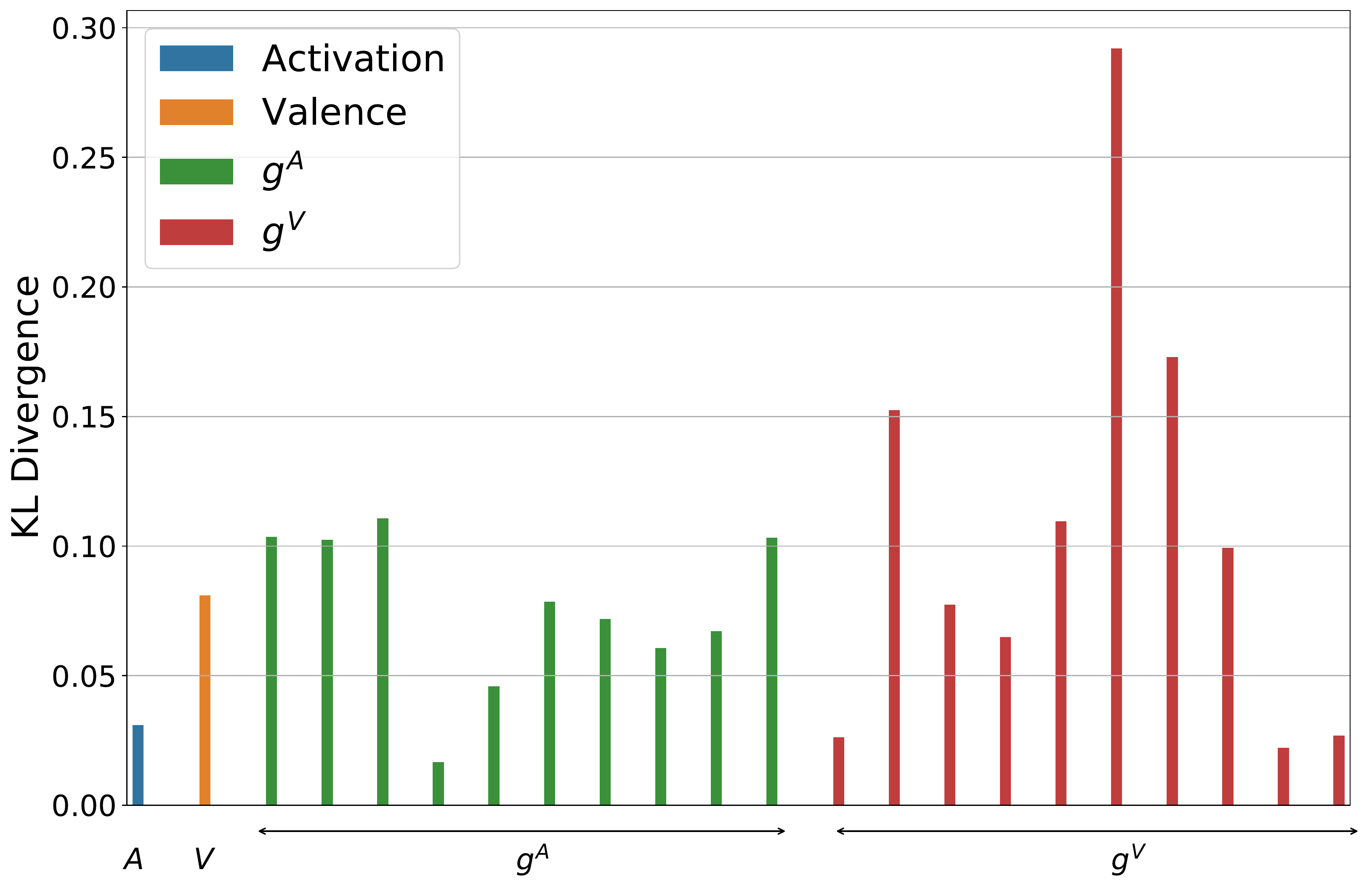}
	\caption{KL divergence $D(P_{f}||Q_{f})$ of each affective feature dimension
	}
	\label{fig:kl_divergence}
\end{figure}

\subsection{Explainability}

We conduct explainability evaluations to better understand contributions of the affective feature dimensions and the proposed model architectures for the affective video summarization. First, we present KL divergence analysis for the affective feature dimensions. Then, we investigate video level summarization performances of the AVSUM models in terms of the cumulative F-score difference metric $\Delta F1_L$ for the \textit{Top-L} human-centric videos.

\subsubsection{Affective Feature Evaluations}

Figure~\ref{fig:kl_divergence} depicts the KL divergence (KLD) of the affective features across distributions gathered on the frames with faces and without faces. A higher KL divergence indicates a bigger discrimination for the distributions of the feature dimension across the with/without face classes. In Figure~\ref{fig:kl_divergence}, the first two KLD values are for the activation and valence attributes, and the later values are color coded for the dimensions of $\mathbf{g}^{A}$ and $\mathbf{g}^{V}$. Note that all the dimensions of the $\mathbf{g}^{A}$, except the fourth, exhibit higher KLD values than the activation attribute, and at certain dimensions KLD is almost 4 times higher than the KLD of the activation. A similar trend can be observed for the valence embedding vector $\mathbf{g}^{V}$, where five dimensions exhibit higher KLD values than the valence attribute, and the largest KLD is extracted for the 6th dimension of the $\mathbf{g}^{V}$. These higher KLD values for the GRU based feature dimensions can be observed as the discriminative cues for the human-centric videos that also contribute to the  performance of the proposed AVSUM architectures.


\begin{figure}[ht]
\centerline{\includegraphics[width=\columnwidth]{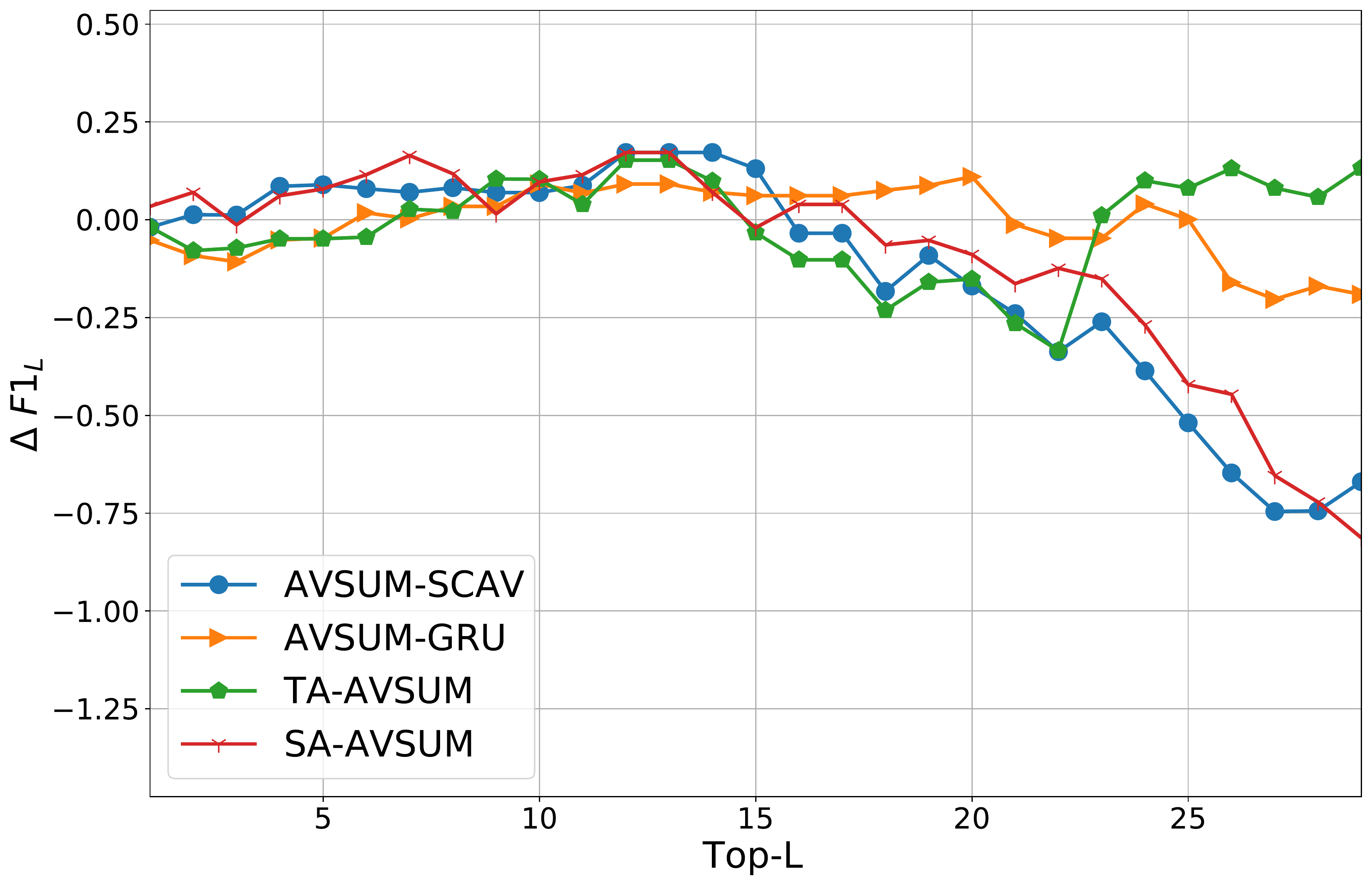}}
\caption{Performance comparison of the AVSUM models with the $\Delta F1_{L}$ metric for the \textit{Top-30} human-centric videos}
\label{fig:delta_f1}
\end{figure}

\subsubsection{Video Level Evaluations}

Figure~\ref{fig:delta_f1} depicts the performance comparison of the AVSUM models with the $\Delta F1_L$ metric for the \textit{Top-30} human-centric videos and yields valuable insights. Note that positive accumulation of $\Delta F1_L$ metric indicates a better performance than the baseline SUM-FCN model. In Figure~\ref{fig:delta_f1}, the AVSUM-SCAV, TA-AVSUM and SA-AVSUM models perform better than the baseline till the \textit{Top-15} human-centric videos, whereas AVSUM-GRU model sustains a higher performance till the \textit{Top-20} human-centric videos. 
Similar trends of the AVSUM-SCAV and SA-AVSUM $\Delta F1_L$ performances can be due to the common late fusion mechanism in these architectures. As $L$ increases, we can assume human-centric characteristic of the videos is getting weaker. Hence both AVSUM-SCAV and SA-AVSUM and as well as AVSUM-GRU models are observed to performed better for the top human-centric videos and their performances start degrading as human-centric characteristic is getting weaker.

\begin{figure*}[htbp]
\centering
\subfigure[$F1$ scores for AVSUM-GRU vs SUM-FCN]{
\includegraphics[width=.95\columnwidth]{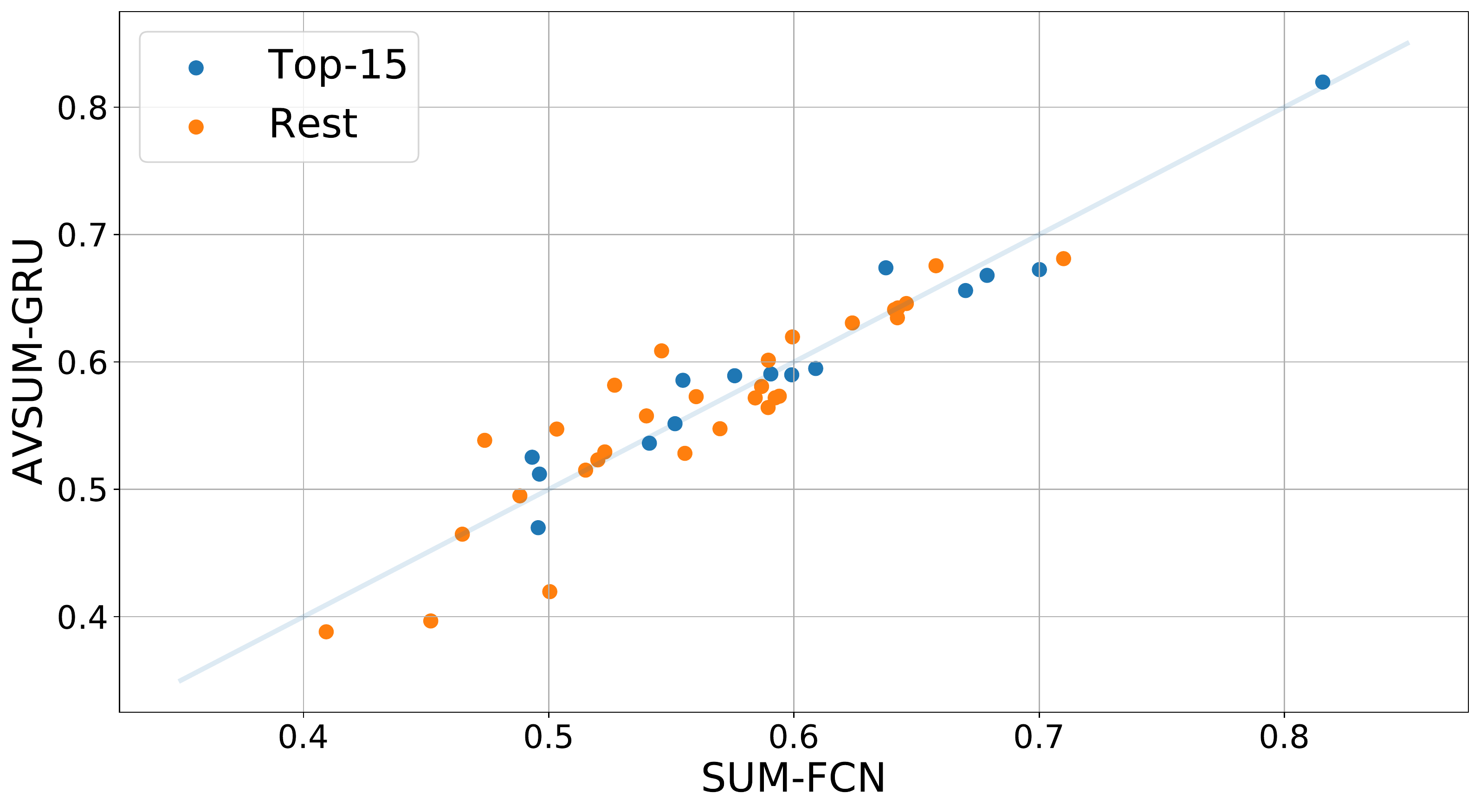}
} \hfill
\subfigure[$R$ scores for AVSUM-GRU vs SUM-FCN]{
\includegraphics[width=.95\columnwidth]{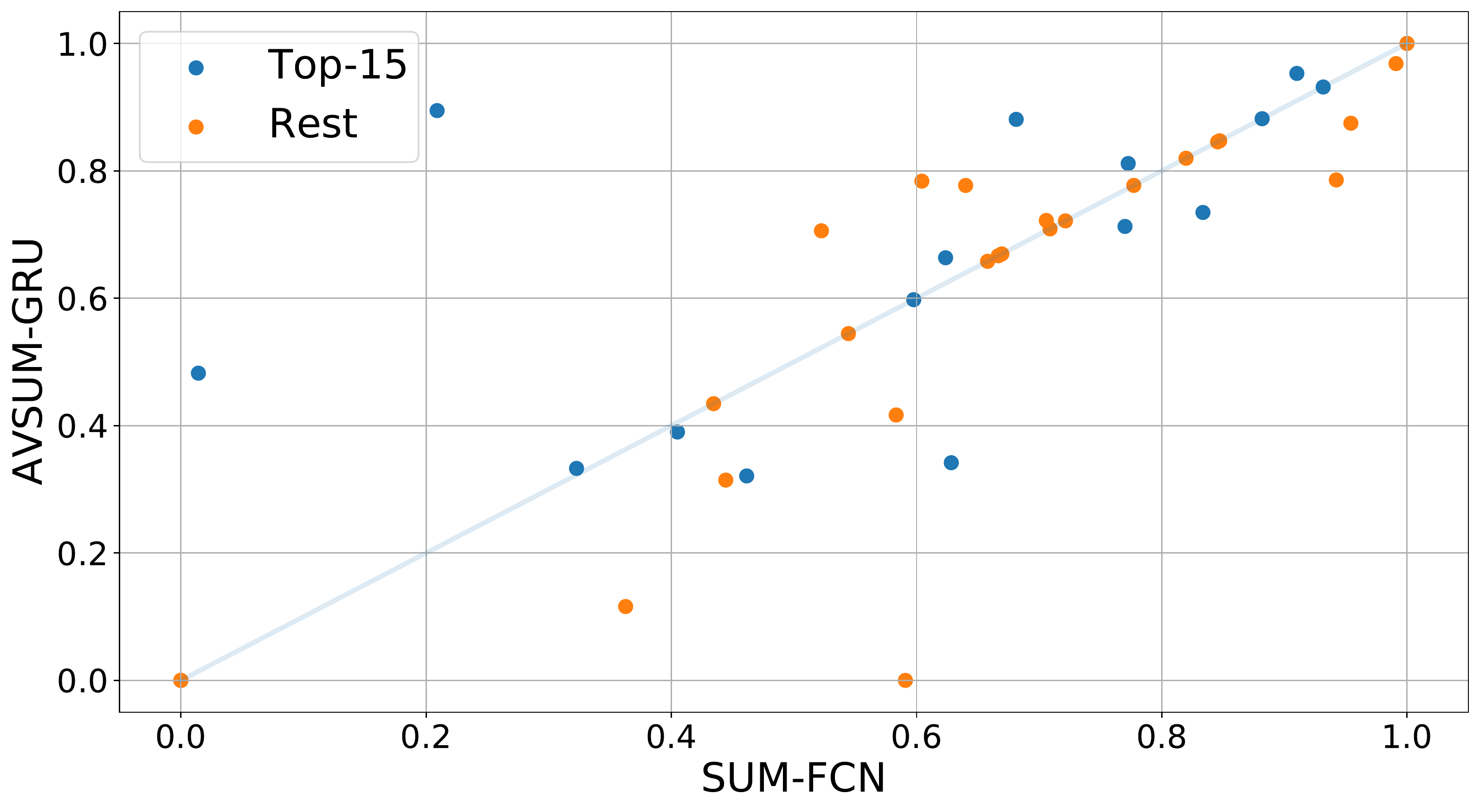}
\label{fig:AVSUM_vs_sum_fcn_face_acc_gru} } \\
\subfigure[$F1$ scores for AVSUM-SCAV vs SUM-FCN]{
\includegraphics[width=.95\columnwidth]{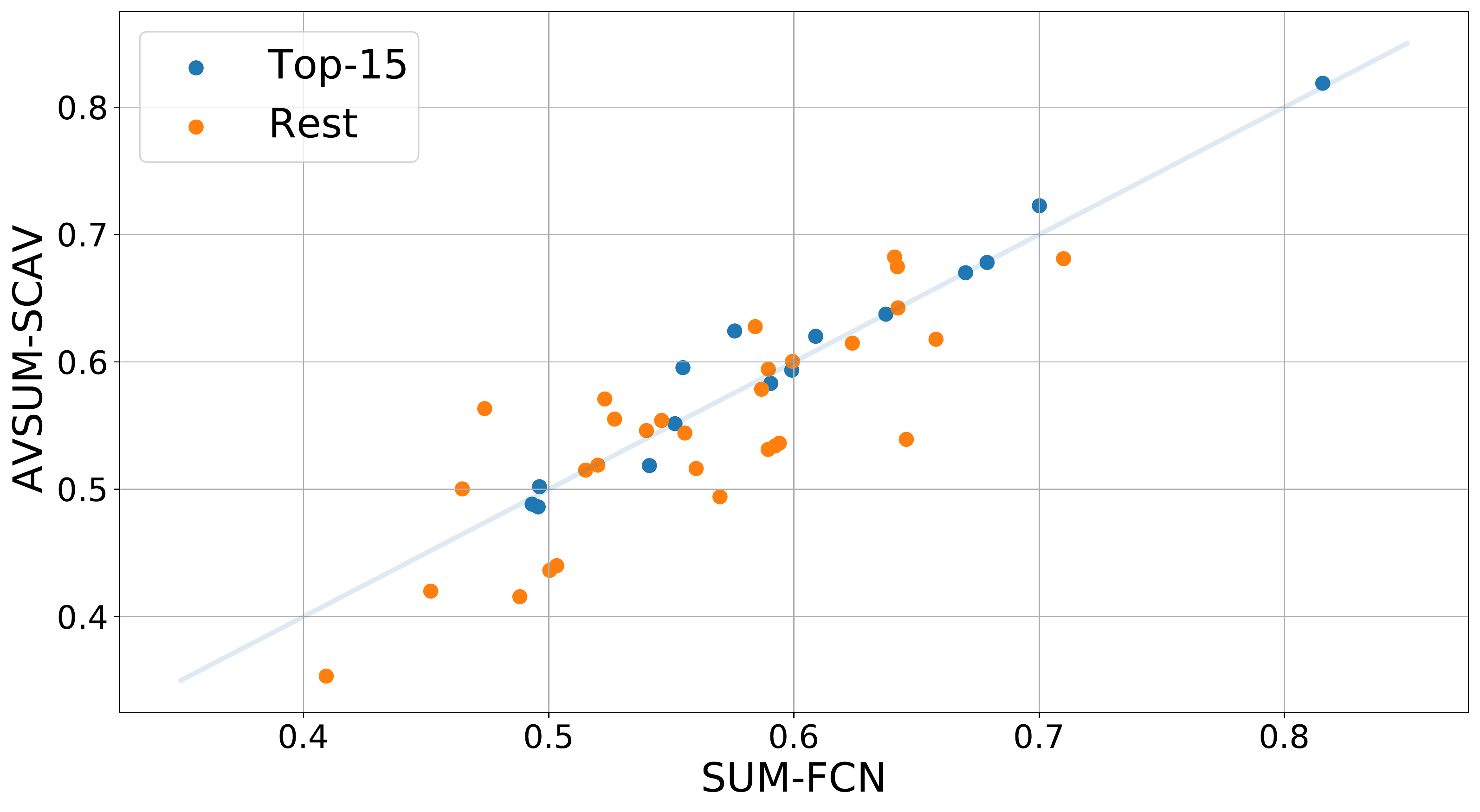}
} \hfill
\subfigure[$R$ scores for AVSUM-SCAV vs SUM-FCN]{
\includegraphics[width=.95\columnwidth]{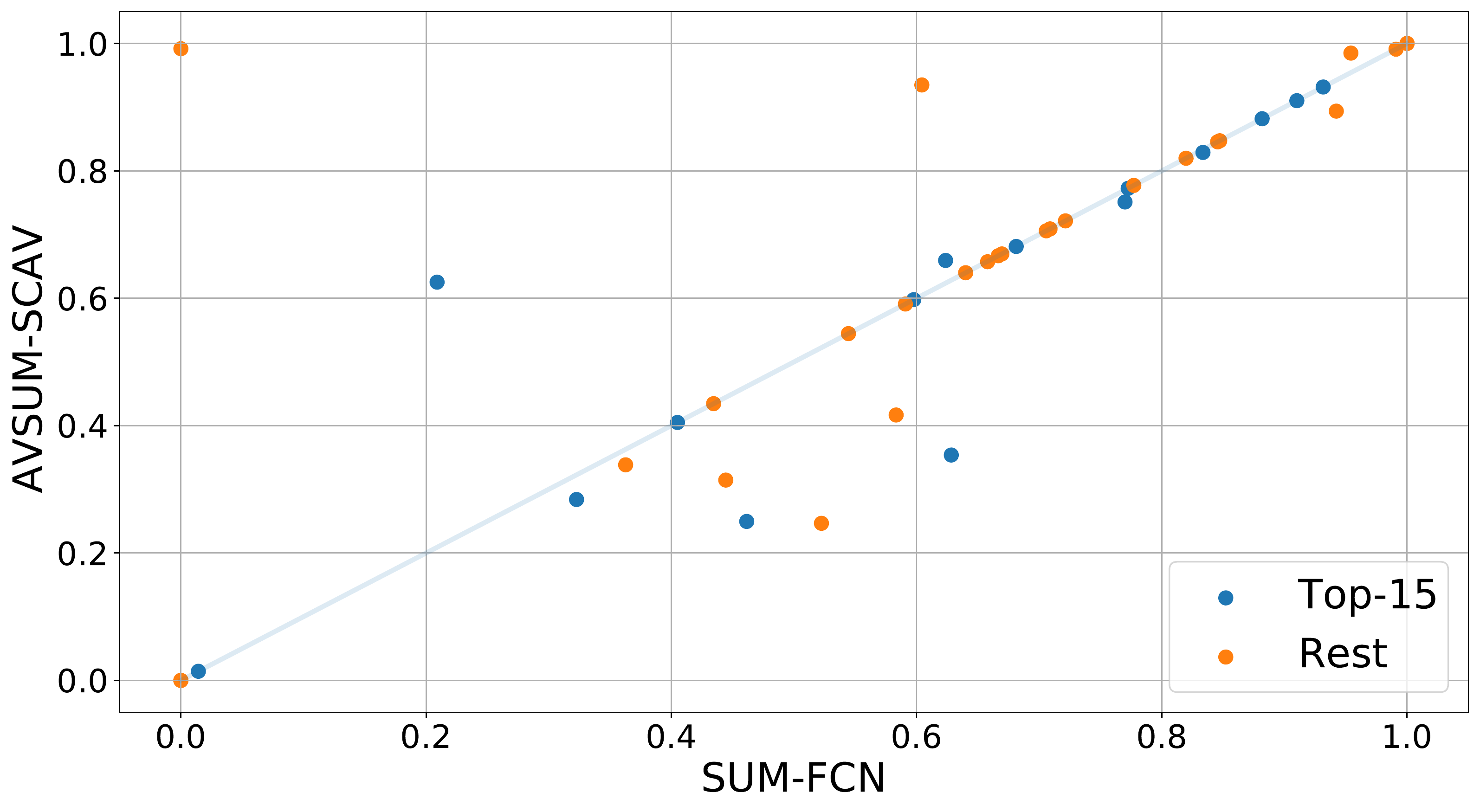}
\label{fig:AVSUM_vs_sum_fcn_face_acc_scav} }  \\
\subfigure[$F1$ scores for TA-AVSUM vs SUM-FCN]{
\includegraphics[width=.95\columnwidth]{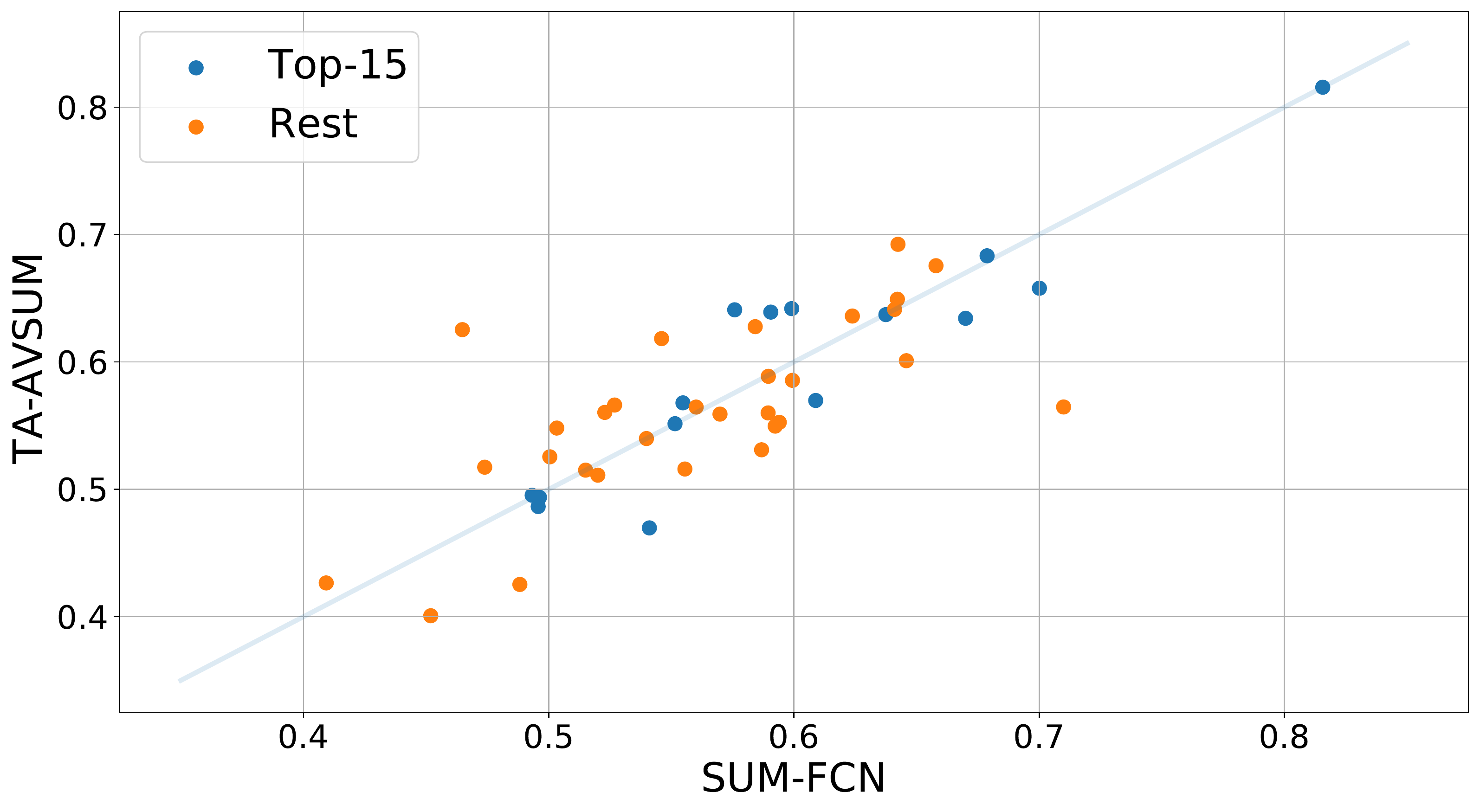}
} \hfill
\subfigure[$R$ scores for TA-AVSUM vs SUM-FCN]{
\includegraphics[width=.95\columnwidth]{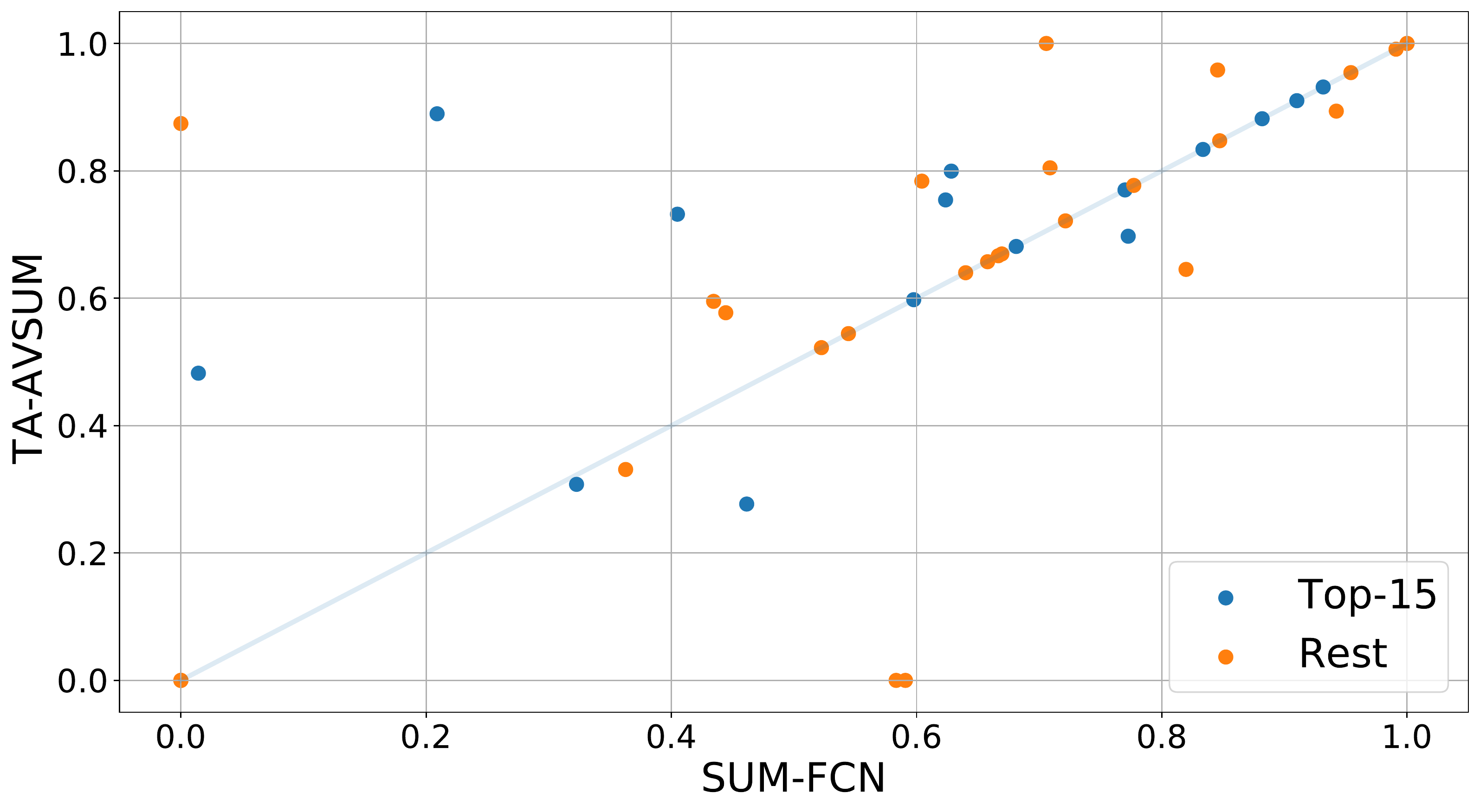}
\label{fig:AVSUM_vs_sum_fcn_face_acc_ta}} \\
\subfigure[$F1$ scores for SA-AVSUM vs SUM-FCN]{
\includegraphics[width=.95\columnwidth]{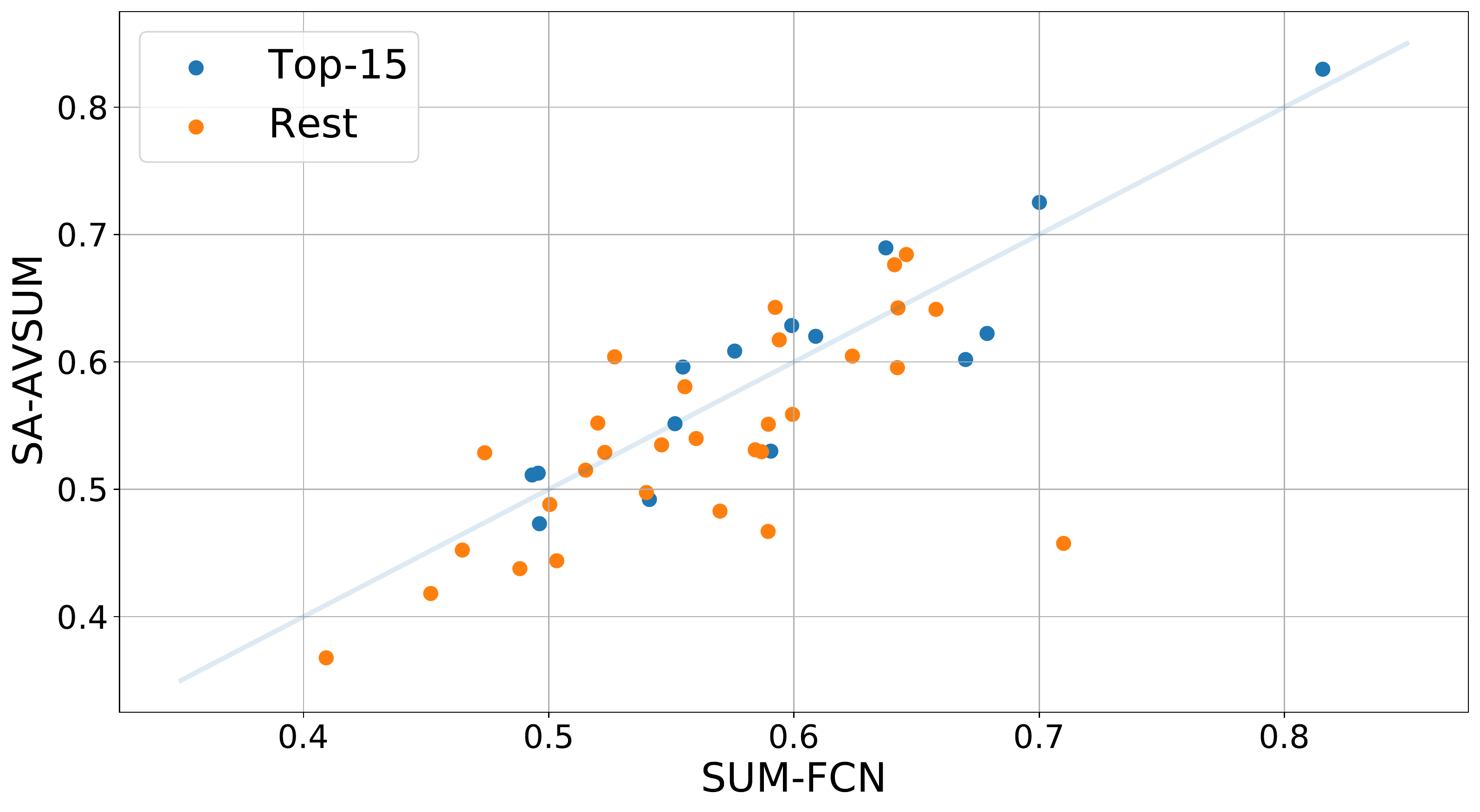}
} \hfill
\subfigure[$R$ scores for SA-AVSUM vs SUM-FCN]{
\includegraphics[width=.95\columnwidth]{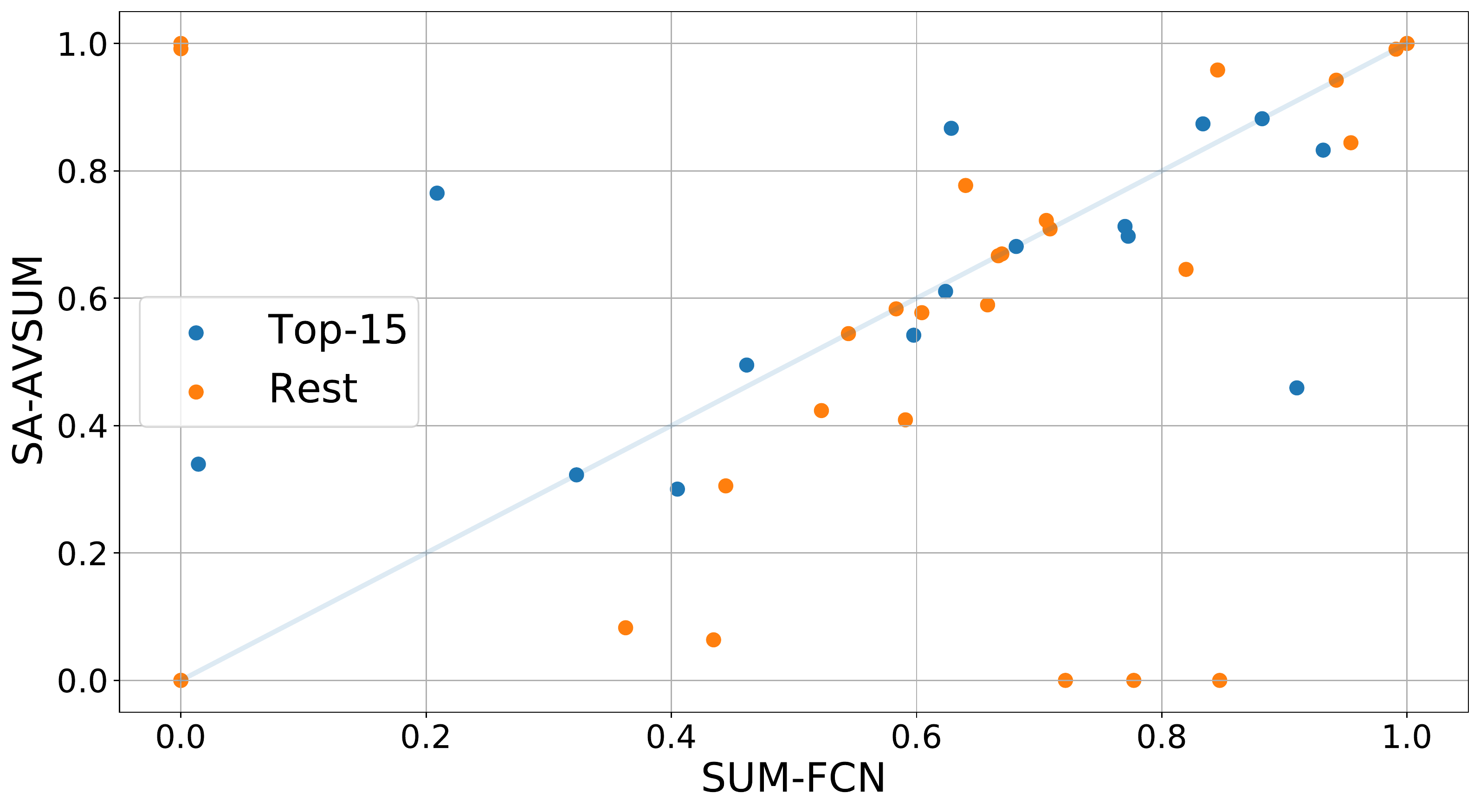}
\label{fig:AVSUM_vs_sum_fcn_face_acc_sa}}

\caption{Video level $F1$ score and $R$ score comparisons of the AVSUM models against the baseline SUM-FCN: $F1$ scores are with Max~$F1$ criterion, $R$ scores are with the Max~$R$ criterion, and the \textit{Top-15} face including videos are color coded in blue.}
\label{fig:AVSUM_vs_sum_fcn}
\end{figure*}

We also investigate video level F-score performances for the proposed AVSUM models. Figure~\ref{fig:AVSUM_vs_sum_fcn} presents scatter plot of the video level $F1$ scores on the left column and video level $R$ scores on the right column, where the diagonal in each figure represents similar performances of the compared models. Furthermore, the \textit{Top-15} human-centric videos are color coded in blue to observe their comparative performances.

Video level F-score performances of the AVSUM-GRU vs SUM-FCN tend to cluster around the diagonal that makes these two models to be most similar in terms of F-score performance. Furthermore, majority of the \textit{Top-15} human-centric videos are on or above the diagonal, which indicates a stronger performance for the human-centric videos. Unlike, F1-score, face recall performance comparison depicted by Figure~\ref{fig:AVSUM_vs_sum_fcn_face_acc_gru} has a scattered behavior providing improvements to some videos while degrading other. However, it is seen that majority of the \textit{Top-15} human-centric videos are affected positively. This observation is in line with the $F1_{15}$ performance of AVSUM-GRU.

Video level F-score performances of the AVSUM-SCAV vs SUM-FCN have a higher deviation from the diagonal. However, almost all the \textit{Top-15} human-centric videos are on or above the diagonal. This indicates a strong performance improvement for the human-centric videos. In terms of face recall on Figure~\ref{fig:AVSUM_vs_sum_fcn_face_acc_scav}, majority of the videos are accumulated around the diagonal, indicating that AVSUM-SCAV and SUM-FCN are most similar in terms of face recall.

Video level F-score performances of the TA-AVSUM and SA-AVSUM models have also high deviation from the diagonal. However, like AVSUM-SCAV majority of the \textit{Top-15} human-centric videos are on or above the diagonal for both. 
Similar to F-score, face recall comparisons of TA-AVSUM and SA-AVSUM have a scattered behavior depicted by Figure~\ref{fig:AVSUM_vs_sum_fcn_face_acc_ta}, and  \ref{fig:AVSUM_vs_sum_fcn_face_acc_sa}. However, different than SA-AVSUM, scattered points accumulated on the positive side for TA-AVSUM, stating a major performance improvement which is inline with its best achieving $R_{15}$ performance for the Max~$R$ criterion.

\section{Conclusion}

In this study, we proposed a new affective information enriched end-to-end video summarization framework for human-centric videos. As a first step, we modeled affective information in terms of AV attributes and GRU embeddings, which were extracted from the CER models. The CER-NET, a CER model achieving state-of-the-art CCC performance, was introduced. We explored the use of affective information with the proposed AVSUM-SCAV and AVSUM-GRU fusion models and attention mechanisms based TA-AVSUM and SA-AVSUM models. Experimental investigations of the proposed models were conducted on the RECOLA and TvSum datasets. 

We observed that with the fusion of affective information, F-score performance of the video summarization on the human-centric videos can be improved. To further analyze the effect of injected features we defined a face recall ($R$) metric and showed that AVSUM-GRU and AVSUM-SCAV models outperform SUM-FCN with more than 1\% increase in face recall $R$. The AVSUM-GRU model has strong performance improvements for the human-centric videos on $F1$ score and face recall $R$, and as well it is the most competitive model with the baseline SUM-FCN. On the other hand, we observed that attention enhanced mechanisms exhibits strong performance gains with the Max~$R$ criterion for the human-centric videos. We should also note that affective GRU embedding features exhibit higher KLD across with-face and without-face frames. 


Comparing the proposed AVSUM models, AVSUM-GRU has a consistent and competitive performance regardless of the model selection criterion at $F1$ and $R$ metrics and has a good balance between the performance improvement at \textit{Top-15} and the performance degradation at the remaining videos. On the other hand, temporal attention based TA-AVSUM performs competitive with the Max~$F1$ criterion and attains strong improvement with the Max~$R$ criterion. The proposed AVSUM models integrate affective information to the summarization architectures and attain important video summarization improvements for the human-centric videos. Compilation of affective human-centric video datasets for video summarization tasks stays as a critical and valuable future study. As a future work, we would like to collect a dataset which is labeled for both video summarization and CER. To extend current study, we would like to investigate multi-modal architectures for both CER and human-centric video summarization components.
  
\IEEEdisplaynontitleabstractindextext

%
\IEEEpeerreviewmaketitle

\ifCLASSOPTIONcaptionsoff
  \newpage
\fi



%
\bibliographystyle{IEEEtran}
\bibliography{ICMR-VASUM.bib}

\begin{thebibliography}{10}
\providecommand{\url}[1]{#1}
\csname url@samestyle\endcsname
\providecommand{\newblock}{\relax}
\providecommand{\bibinfo}[2]{#2}
\providecommand{\BIBentrySTDinterwordspacing}{\spaceskip=0pt\relax}
\providecommand{\BIBentryALTinterwordstretchfactor}{4}
\providecommand{\BIBentryALTinterwordspacing}{\spaceskip=\fontdimen2\font plus
\BIBentryALTinterwordstretchfactor\fontdimen3\font minus
  \fontdimen4\font\relax}
\providecommand{\BIBforeignlanguage}[2]{{%
\expandafter\ifx\csname l@#1\endcsname\relax
\typeout{** WARNING: IEEEtran.bst: No hyphenation pattern has been}%
\typeout{** loaded for the language `#1'. Using the pattern for}%
\typeout{** the default language instead.}%
\else
\language=\csname l@#1\endcsname
\fi
#2}}
\providecommand{\BIBdecl}{\relax}
\BIBdecl

\bibitem{kopruCERMTL2020}
B.~Köprü and E.~Erzin, ``Multimodal continuous emotion recognition using deep
  multi-task learning with correlation loss,'' \emph{arXiv:2011.00876}, 2020.

\bibitem{videoSumSurvey1}
E.~Apostolidis, E.~Adamantidou, A.~I. Metsai, V.~Mezaris, and I.~Patras,
  ``Video summarization using deep neural networks: A survey,''
  \emph{arXiv:2101.06072v1}, 2021.

\bibitem{emotionDecisionMaking2015}
J.~S. Lerner, Y.~Li, P.~Valdesolo, and K.~S. Kassam, ``{Emotion and decision
  making},'' \emph{Annual Review of Psychology}, vol.~66, pp. 799--823, jan
  2015.

\bibitem{threeDimensionalEmotion12}
J.~A. Russell and A.~Mehrabian, ``Evidence for a three-factor theory of
  emotions,'' \emph{Journal of Research in Personality}, vol.~11, no.~3, pp.
  273--294, 1977.

\bibitem{threeDimensionalEmotioin54}
H.~Schlosberg, ``Three dimensions of emotion,'' \emph{Psychological Review},
  vol.~61, no.~2, pp. 81--88, 1954.

\bibitem{speechCER16}
G.~Trigeorgis, F.~Ringeval, R.~Brueckner, E.~Marchi, M.~A. Nicolaou,
  B.~Schuller, and S.~Zafeiriou, ``Adieu features? end-to-end speech emotion
  recognition using a deep convolutional recurrent network,'' in \emph{ICASSP
  2016, IEEE International Conference on Acoustics, Speech and Signal
  Processing}, 2016, pp. 5200--5204.

\bibitem{speechCER17}
M.~Neumann and N.~T. Vu, ``Attentive convolutional neural network based speech
  emotion recognition: A study on the impact of input features, signal length,
  and acted speech,'' in \emph{INTERSPEECH 2017, Annual Conference of the
  International Speech Communication Association}, 2017, pp. 1263--1267.

\bibitem{speechCER19}
Y.~Li, T.~Zhao, and T.~Kawahara, ``Improved end-to-end speech emotion
  recognition using self attention mechanism and multitask learning,'' in
  \emph{INTERSPEECH 2019, Annual Conference of the International Speech
  Communication Association}, 2019, pp. 2803--2807.

\bibitem{videoCER1}
S.~Minaee and A.~Abdolrashidi, ``{Deep-Emotion}: Facial expression recognition
  using attentional convolutional network,'' \emph{arXiv:1902.01019}, 2019.

\bibitem{videoCER2}
D.~Aspandi, A.~Mallol-Ragolta, B.~Schuller, and X.~Binefa, ``{Adversarial-based
  neural networks for affect estimations in the wild},''
  \emph{arXiv:2002.00883}, 2020.

\bibitem{videoCER3}
D.~Kollias, M.~A. Nicolaou, I.~Kotsia, G.~Zhao, and S.~Zafeiriou, ``Recognition
  of affect in the wild using deep neural networks,'' in \emph{IEEE Computer
  Society Conference on Computer Vision and Pattern Recognition Workshops},
  2017, pp. 1972--1979.

\bibitem{frameimportanceToGroundTruth2016}
K.~Zhang, W.-L. Chao, F.~Sha, and K.~Grauman, ``Video summarization with long
  short-term memory,'' in \emph{ECCV 2016, European Conference on Computer
  Vision}.\hskip 1em plus 0.5em minus 0.4em\relax Springer International
  Publishing, 2016, pp. 766--782.

\bibitem{earlyVideoSum1}
Y.-F. Ma, L.~Lu, H.-J. Zhang, and M.~Li, ``A user attention model for video
  summarization,'' in \emph{MULTIMEDIA '02, Tenth ACM International Conference
  on Multimedia}, 2002, pp. 533--542.

\bibitem{earlyVideoSum5}
G.~Kim and E.~P. Xing, ``{Reconstructing storyline graphs for image
  recommendation from web community photos},'' in \emph{CVPR 2014, IEEE
  Conference on Computer Vision and Pattern Recognition}, 2014, pp. 3882--3889.

\bibitem{earlyVideoSum6}
P.~Mundur, Y.~Rao, and Y.~Yesha, ``Keyframe-based video summarization using
  {Delaunay} clustering,'' \emph{International Journal on Digital Libraries},
  vol.~6, no.~2, pp. 219--232, apr 2006.

\bibitem{earlyVideoSum2ColorHist}
C.~W. Ngo, T.~C. Pong, H.~J. Zhang, and R.~T. Chin, ``{Motion-based video
  representation for scene change detection},'' in \emph{ICPR 2000,
  International Conference on Pattern Recognition}, vol.~15, no.~1, 2000, pp.
  827--830.

\bibitem{earlVideoSum3ColorHist}
C.~W. Ngo, Y.~F. Ma, and H.~J. Zhang, ``{Automatic video summarization by graph
  modeling},'' in \emph{ICCV 2003, IEEE International Conference on Computer
  Vision}, vol.~1, 2003, pp. 104--109.

\bibitem{earlyVideoSum4MI}
Z.~Cernekova, I.~Pitas, and C.~Nikou, ``{Information theory-based shot cut/fade
  detection and video summarization},'' \emph{IEEE Transactions on Circuits and
  Systems for Video Technology}, vol.~16, no.~1, pp. 82--91, jan 2006.

\bibitem{unsupervisedVideoSum1}
B.~Mahasseni, M.~Lam, and S.~Todorovic, ``{Unsupervised video summarization
  with adversarial LSTM networks},'' in \emph{CVPR 2017, 30th IEEE Conference
  on Computer Vision and Pattern Recognition}, 2017, pp. 2982--2991.

\bibitem{unsupervisedVideoSumAttention1}
E.~Apostolidis, E.~Adamantidou, A.~I. Metsai, V.~Mezaris, and I.~Patras,
  ``Unsupervised video summarization via attention-driven adversarial
  learning,'' in \emph{MultiMedia Modeling}, Y.~M. Ro, W.-H. Cheng, J.~Kim,
  W.-T. Chu, P.~Cui, J.-W. Choi, M.-C. Hu, and W.~De~Neve, Eds.\hskip 1em plus
  0.5em minus 0.4em\relax Springer International Publishing, 2020, pp.
  492--504.

\bibitem{unsupervisedVideoSumAttention2}
X.~He, Y.~Hua, T.~Song, Z.~Zhang, Z.~Xue, R.~Ma, N.~Robertson, and H.~Guan,
  ``Unsupervised video summarization with attentive conditional generative
  adversarial networks,'' in \emph{Multimedia 2019, 27th ACM International
  Conference on Multimedia}, 2019, p. 2296–2304.

\bibitem{videoSumFCSN2018}
M.~Rochan, L.~Ye, and Y.~Wang, ``{Video summarization using fully convolutional
  sequence networks},'' in \emph{ECCV 2018, European Conference on Computer
  Vision}, 2018, pp. 358--374.

\bibitem{videoSummAttentionEncoderDecoder2020}
Z.~Ji, K.~Xiong, Y.~Pang, and X.~Li, ``Video summarization with attention-based
  encoder–decoder networks,'' \emph{IEEE Transactions on Circuits and Systems
  for Video Technology}, vol.~30, no.~6, pp. 1709--1717, 2020.

\bibitem{supervisedVideoSumRW1}
P.~Li, Q.~Ye, L.~Zhang, L.~Yuan, X.~Xu, and L.~Shao, ``Exploring global diverse
  attention via pairwise temporal relation for video summarization,''
  \emph{Pattern Recognition}, vol. 111, p. 107677, 2021.

\bibitem{affectiveSumFromViewer1}
H.~Joho, J.~Staiano, N.~Sebe, and J.~M. Jose, ``{Looking at the viewer:
  Analysing facial activity to detect personal highlights of multimedia
  contents},'' \emph{Multimedia Tools and Applications}, vol.~51, no.~2, pp.
  505--523, jan 2011.

\bibitem{affectiveSumFromViewer3}
A.~G. Money and H.~Agius, ``{Analysing user physiological responses for
  affective video summarisation},'' \emph{Displays}, vol.~30, no.~2, pp.
  59--70, apr 2009.

\bibitem{discreteER1}
J.~Zhao, X.~Mao, and L.~Chen, ``{Speech emotion recognition using deep 1D \& 2D
  CNN LSTM networks},'' \emph{Biomedical Signal Processing and Control},
  vol.~47, pp. 312--323, jan 2019.

\bibitem{discreteER2}
S.~Mirsamadi, E.~Barsoum, and C.~Zhang, ``{Automatic speech emotion recognition
  using recurrent neural networks with local attention},'' in \emph{ICASSP
  2017, IEEE International Conference on Acoustics, Speech and Signal
  Processing}, 2017, pp. 2227--2231.

\bibitem{discreteER3}
H.~Meng, T.~Yan, F.~Yuan, and H.~Wei, ``Speech emotion recognition from {3D}
  log-mel spectrograms with deep learning network,'' \emph{IEEE Access},
  vol.~7, pp. 125\,868--125\,881, 2019.

\bibitem{contER_CCC1}
M.~Schmitt, N.~Cummins, and B.~W. Schuller, ``Continuous emotion recognition in
  speech — do we need recurrence?'' in \emph{INTERSPEECH 2019, Annual
  Conference of the International Speech Communication Association}.\hskip 1em
  plus 0.5em minus 0.4em\relax ISCA, 2019, pp. 2808--2812.

\bibitem{multimodalCERRW1}
P.~Tzirakis, J.~Chen, S.~Zafeiriou, and B.~Schuller, ``End-to-end multimodal
  affect recognition in real-world environments,'' \emph{Information Fusion},
  vol.~68, pp. 46--53, apr 2021.

\bibitem{emotionSummarization1}
B.~Xu, Y.~Fu, Y.~G. Jiang, B.~Li, and L.~Sigal, ``{Heterogeneous knowledge
  transfer in video emotion recognition, attribution and summarization},''
  \emph{IEEE Transactions on Affective Computing}, vol.~9, no.~2, pp. 255--270,
  apr 2018.

\bibitem{emotionSummarization2}
G.~Tu, Y.~Fu, B.~Li, J.~Gao, Y.~G. Jiang, and X.~Xue, ``{A Multi-Task Neural
  Approach for Emotion Attribution, Classification, and Summarization},''
  \emph{IEEE Transactions on Multimedia}, vol.~22, no.~1, pp. 148--159, 2020.

\bibitem{unsupervisedVideoSumRW1}
Y.~Jung, D.~Cho, D.~Kim, S.~Woo, and I.~S. Kweon, ``Discriminative feature
  learning for unsupervised video summarization,'' in \emph{33rd AAAI
  Conference on Artificial Intelligence}, vol.~33, no.~01, 2019, pp.
  8537--8544.

\bibitem{unsupervisedVideoSumRW3}
B.~Zhao, X.~Li, and X.~Lu, ``Property-constrained dual learning for video
  summarization,'' \emph{IEEE Transactions on Neural Networks and Learning
  Systems}, vol.~31, no.~10, pp. 3989--4000, oct 2020.

\bibitem{unsupervisedVideoSumRW2}
K.~Zhou, Y.~Qiao, and T.~Xiang, ``Deep reinforcement learning for unsupervised
  video summarization with diversity-representativeness reward,'' in \emph{AAAI
  Conference on Artificial Intelligence}, vol.~32, no.~1, 2018.

\bibitem{encodeDecoderVSum2019}
Y.~Yuan, H.~Li, and Q.~Wang, ``Spatiotemporal modeling for video summarization
  using convolutional recurrent neural network,'' \emph{IEEE Access}, vol.~7,
  pp. 64\,676--64\,685, 2019.

\bibitem{attention1}
D.~Bahdanau, K.~H. Cho, and Y.~Bengio, ``{Neural machine translation by jointly
  learning to align and translate},'' in \emph{ICLR 2015, 3rd International
  Conference on Learning Representations}, 2015.

\bibitem{attention2}
A.~Vaswani, N.~Shazeer, N.~Parmar, J.~Uszkoreit, L.~Jones, A.~N. Gomez,
  L.~Kaiser, and I.~Polosukhin, ``Attention is all you need,''
  \emph{arXiv:1706.03762}, 2017.

\bibitem{videoSummAttentionEncoderDecoder2021}
Z.~Ji, F.~Jiao, Y.~Pang, and L.~Shao, ``{Deep attentive and semantic preserving
  video summarization},'' \emph{Neurocomputing}, vol. 405, pp. 200--207, sep
  2020.

\bibitem{tvSum2018}
Y.~Song, J.~Vallmitjana, A.~Stent, and A.~Jaimes, ``{TVSum}: Summarizing web
  videos using titles,'' in \emph{CVPR 2015, IEEE Conference on Computer Vision
  and Pattern Recognition}, 2015, pp. 5179--5187.

\bibitem{googlenet2015}
C.~Szegedy, W.~Liu, Y.~Jia, P.~Sermanet, S.~Reed, D.~Anguelov, D.~Erhan,
  V.~Vanhoucke, and A.~Rabinovich, ``Going deeper with convolutions,'' in
  \emph{CVPR 2015, IEEE Conference on Computer Vision and Pattern Recognition},
  2015, pp. 1--9.

\bibitem{fcn2014}
J.~Long, E.~Shelhamer, and T.~Darrell, ``{Fully Convolutional Networks for
  Semantic Segmentation},'' \emph{IEEE Transactions on Pattern Analysis and
  Machine Intelligence}, vol.~39, no.~4, pp. 640--651, nov 2014.

\bibitem{squeexeExcitationNet}
J.~Hu, L.~Shen, and G.~Sun, ``Squeeze-and-excitation networks,'' in \emph{CVPR
  2018, IEEE Conference on Computer Vision and Pattern Recognition}, 2018, pp.
  7132--7141.

\bibitem{recola2013}
F.~Ringeval, A.~Sonderegger, J.~Sauer, and D.~Lalanne, ``Introducing the
  {RECOLA} multimodal corpus of remote collaborative and affective
  interactions,'' in \emph{FG 2013, 10th IEEE International Conference and
  Workshops on Automatic Face and Gesture Recognition}, 2013.

\bibitem{hog_face_detector_dlib}
D.~E. King, ``Dlib-ml: A machine learning toolkit,'' \emph{Journal of Machine
  Learning Research}, vol.~10, no.~60, pp. 1755--1758, 2009.

\bibitem{endEndVisualNetCER}
P.~Tzirakis, G.~Trigeorgis, M.~A. Nicolaou, B.~W. Schuller, and S.~Zafeiriou,
  ``End-to-end multimodal emotion recognition using deep neural networks,''
  \emph{IEEE Journal of Selected Topics in Signal Processing}, vol.~11, no.~8,
  pp. 1301--1309, 2017.

\end{thebibliography}

%
\begin{IEEEbiography}[{\includegraphics[width=1in,height=1.25in,clip,keepaspectratio]{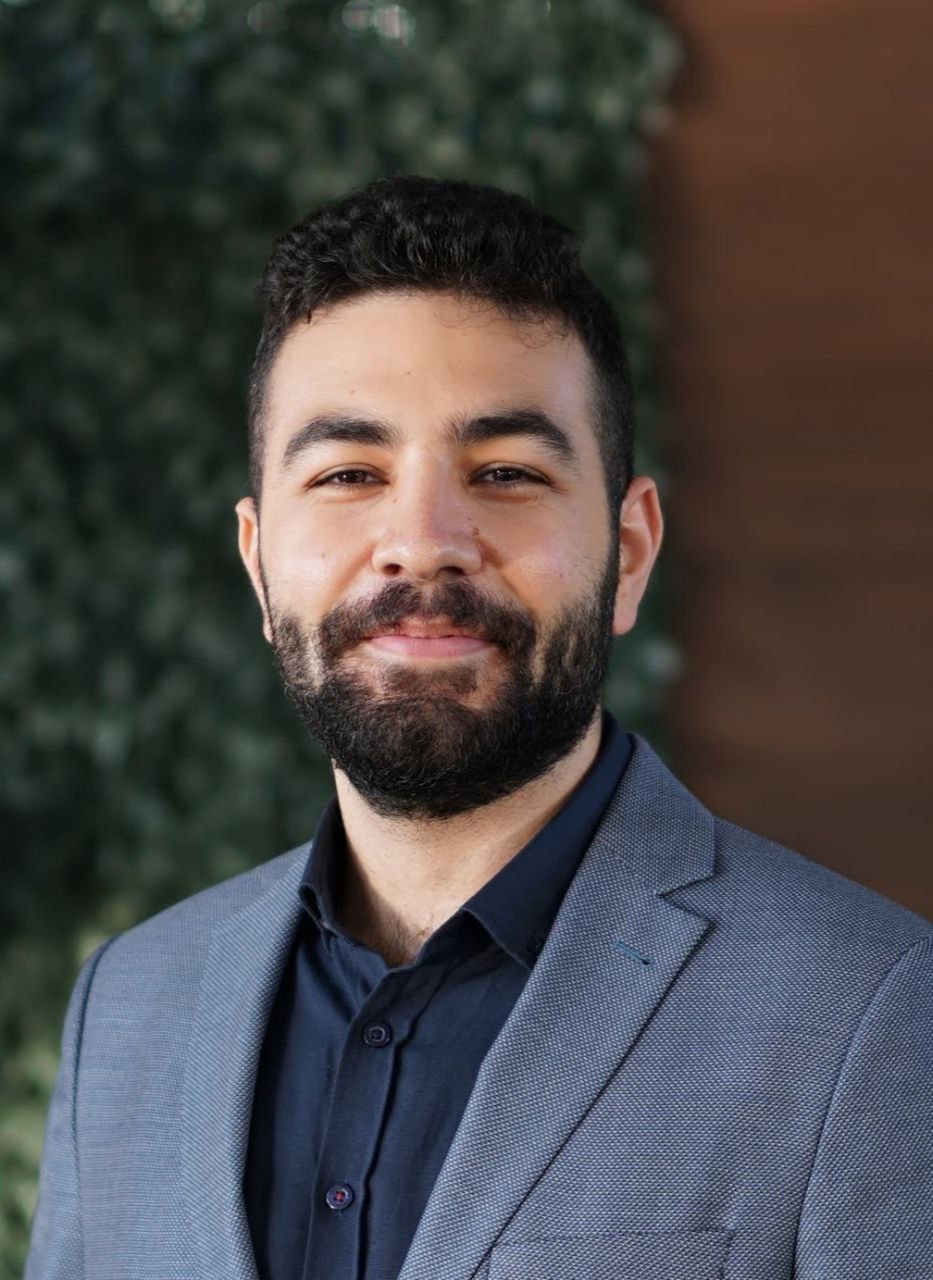}}]{Berkay Köprü}

is a Ph.D. student at Koc University, Istanbul, Turkey. He received his M.Sc. degree and B.Sc. degree from Technical University of Munich, Munich, Germany in 2017, and Bilkent University, Ankara, Turkey in 2014 respectively. His research interest include human-computer interaction, computer vision and affective computing.
\end{IEEEbiography}

\begin{IEEEbiography}[{\includegraphics[width=1in,height=1.25in,clip,keepaspectratio]{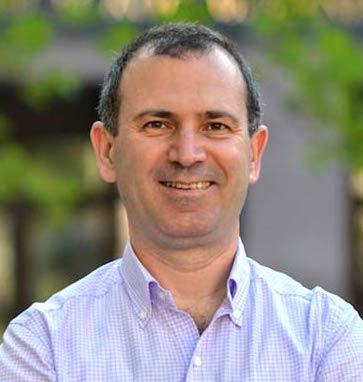}}]{Engin Erzin}
(S’88-M’96-SM’06) received his Ph.D. degree, M.Sc. degree, and B.Sc. degree from the Bilkent University, Ankara, Turkey, in 1995, 1992 and 1990, respectively, all in Electrical Engineering. During 1995-1996, he was a postdoctoral fellow in Signal Compression Laboratory, University of California, Santa Barbara. He joined Lucent Technologies in September 1996, and he was with the Consumer Products for one year as a Member of Technical Staff of the Global Wireless Products Group. From 1997 to 2001, he was with the Speech and Audio Technology Group of the Network Wireless Systems. Since January 2001, he has been with the  Electrical \& Electronics Engineering and Computer Engineering Departments of Koc University, Istanbul, Turkey.  Engin Erzin is currently a member of the IEEE Speech and Language Processing Technical Committee and Associate Editor for the IEEE Transactions on Multimedia, having previously served as Associate Editor of the IEEE Transactions on Audio, Speech \& Language Processing (2010-2014). His research interests include speech-audio-visual signal processing, affective computing, human-computer interaction and machine learning.
\end{IEEEbiography}
%




\end{document}